\documentclass[10pt,twocolumn,letterpaper]{article}

\usepackage{iccv}
\usepackage{times}
\usepackage{epsfig}
\usepackage{graphicx}
\usepackage{amsmath}
\usepackage{multirow}
\usepackage{amssymb}
\usepackage{pifont}
\usepackage{mathtools}

\DeclarePairedDelimiter\floor{\lfloor}{\rfloor}
\usepackage{float}
\usepackage[breaklinks=true,bookmarks=false]{hyperref}

\iccvfinalcopy 

\def\iccvPaperID{****} 

\begin{document}

\title{Bringing Background into the Foreground: \\Making All Classes Equal in Weakly-supervised Video Semantic Segmentation}
 \author{Fatemeh Sadat Saleh$^{1,2}$, Mohammad Sadegh Aliakbarian$^{1,2}$, Mathieu Salzmann$^{3}$, \\Lars Petersson$^{1,2}$, Jose M. Alvarez$^{1,2}$\\
$^{1}$Australian National University, $^{2}$Smart Vision Systems, CSIRO, $^{3}$CVLab, EPFL\\
 {\tt\small firstname.lastname@data61.csiro.au, mathieu.salzmann@epfl.ch}
}

\maketitle
\begin{abstract}
Pixel-level annotations are expensive and time-consuming to obtain. Hence, weak supervision using only image tags could have a significant impact in semantic segmentation. Recent years have seen great progress in weakly-supervised semantic segmentation, whether from a single image or from videos. However, most existing methods are designed to handle a single background class. In practical applications, such as autonomous navigation, it is often crucial to reason about multiple background classes. In this paper, we introduce an approach to doing so by making use of classifier heatmaps. We then develop a two-stream deep architecture that jointly leverages appearance and motion, and design a loss based on our heatmaps to train it. Our experiments demonstrate the benefits of our classifier heatmaps and of our two-stream architecture on challenging urban scene datasets and on the YouTube-Objects benchmark, where we obtain state-of-the-art results.

\end{abstract}

\section{Introduction}
Video semantic segmentation, i.e., the task of assigning a semantic label to every pixel in video frames, is crucial for the success of many computer vision applications, such as  video summarization and autonomous navigation. In this context, fully-supervised methods~\cite{featurespace,voxel2voxel,clockwork,tripathi,predective} have made great progress, particularly with the advent of deep learning. These methods, however, inherently rely on having access to large amounts of training videos with pixel-level ground-truth annotations in every frame. Unfortunately, such annotations are highly time-consuming and expensive to obtain, and generating realistic synthetic data to obtain annotations~\cite{synthia,playgame} is a challenging task in itself.
While semi-supervised techniques~\cite{supervoxel,viaflow,viastroke} mitigate this issue by leveraging partial annotations, they still require some pixel-level ground-truth.

\begin{figure}[t!]
\centering
\includegraphics[width=0.45\textwidth]{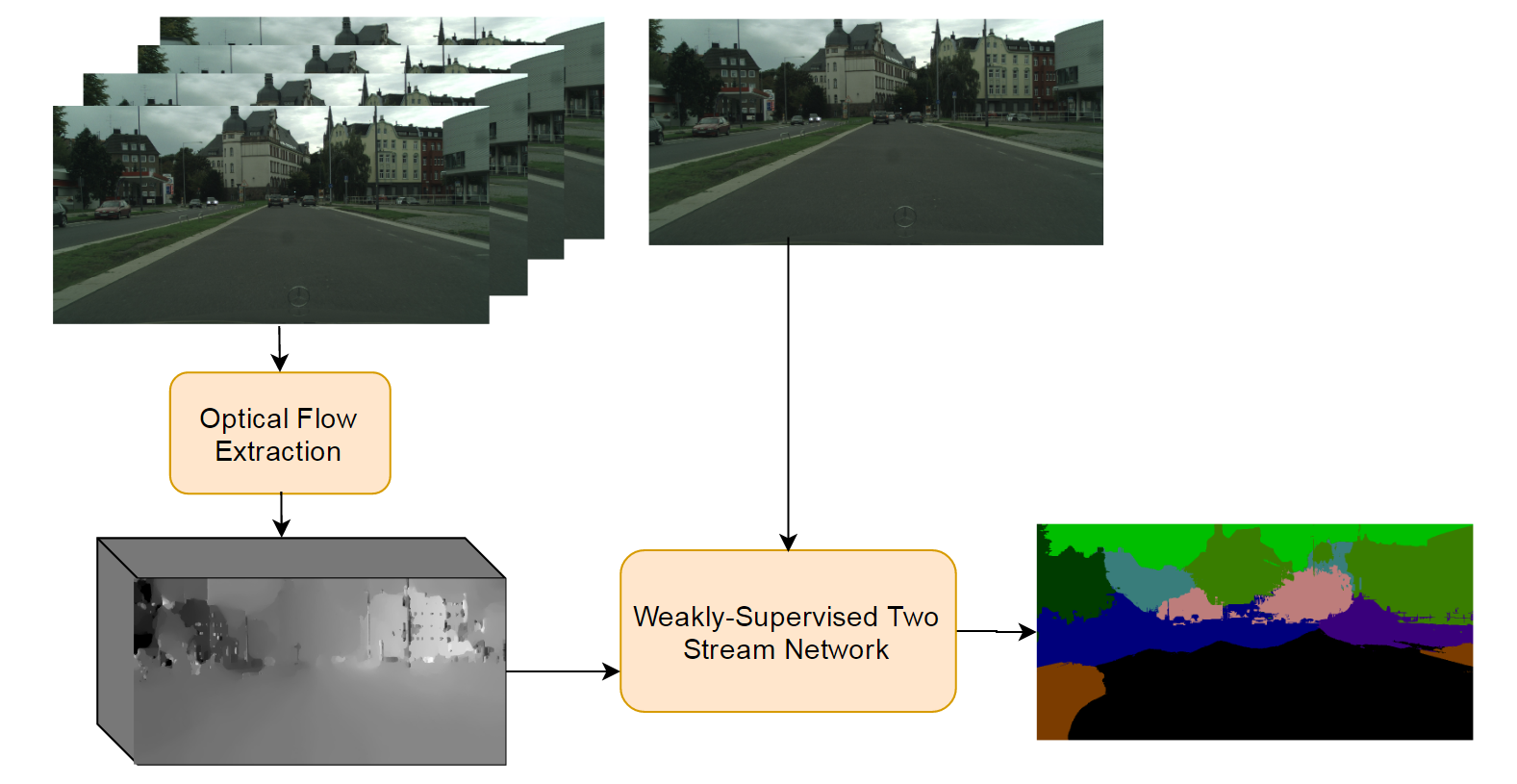}
\caption{\textbf{Overview of our framework}. Given only video-level tags, our weakly-supervised video semantic segmentation network jointly leverages classifier heatmaps and motion information to model both multiple foreground classes and multiple background classes. This is in contrast with most methods that focus on foreground classes only, thus being  inapplicable to scenarios where differentiating background classes is crucial, such as in autonomous driving.}
\label{fig:method}
\end{figure}

By contrast, weakly-supervised semantic segmentation methods~\cite{webscale,multi-class,viadetection,discriminative,domain_adapt,obj_track,movingobject,fast,saleh,sec,ccnn,stc,whatspoint,weaklyandsemi} rely only on tags. When working with still images~\cite{saleh,sec,ccnn,stc,whatspoint,weaklyandsemi}, tags are typically assumed to be available in each image, whereas for video-based segmentation~\cite{webscale,multi-class,viadetection,discriminative,domain_adapt,obj_track,movingobject,fast}, tags correspond to entire videos or video snippets.
While recent years have seen great progress in weakly-supervised semantic segmentation, most existing methods, whether image- or video-based, have a major drawback: They focus on foreground object classes and treat the background as one single entity. However, having detailed information about the different background classes is crucial in many practical scenarios, such as autonomous driving, where one needs to differentiate the road from a grass field. 

In this paper, we introduce an approach to weakly-supervised video semantic segmentation that treats all classes, foreground and background ones, equally (see Fig~\ref{fig:method}). To this end, we propose to rely on class-dependent heatmaps obtained from classifiers trained for image-level recognition, i.e., requiring no pixel-level annotations. These classifier heatmaps provide us with valuable information about the location of instances/regions of each class. We therefore introduce a weakly-supervised loss function that let us exploit them in a deep architecture.

In particular, we develop a two-stream deep network that jointly leverages appearance and motion. Our network fuses these two complementary sources of information in two different ways: A trainable early fusion, which puts in correspondence the spatial and temporal information and learns to combine it into a spatio-temporal stream, and a late fusion further leveraging the valuable semantic information of the spatial stream to merge it with the spatio-temporal one for final prediction. Altogether, our approach constitutes the first end-to-end framework for weakly-supervised semantic segmentation to handle both multiple foreground and background classes.

To the best of our knowledge, only two weakly-supervised video semantic segmentation approaches~\cite{multi-class,co_parsing} can potentially handle multiple background classes. However,~\cite{multi-class} relies on a simple similarity measure between handcrafted features, and thus does not translate well to complex scenes where multiple instances of the same class have significantly different appearances. 
While~\cite{co_parsing} relies on more robust, pre-trained deep learning features, it exploits additional, pixel-wise annotations to train a fully-convolutional network for scene/object classification. Furthermore, none of these two methods offer an end-to-end learning approach, which has proven key to the success of many other computer vision tasks.

Our experiments demonstrate the benefits of our approach in several scenarios. First, it yields accurate segmentations on challenging outdoor scenes, such as those depicted by the CamVid~\cite{camvid} and CityScapes~\cite{Cityscapes} datasets, for which methods modeling foreground classes only do not apply. Furthermore, it outperforms the state-of-the-art methods that, as us, rely only on video-level tags on the standard YouTube Object~\cite{youtube} dataset.

\section{Related Work}
Over the years, many approaches have tackled the problem of video semantic segmentation. In particular, much research has been done in the context of fully-supervised semantic segmentation, including methods based on CNNs~\cite{clockwork,voxel2voxel,predective} and on graphical models~\cite{featurespace,activeinference,tripathi}. Here, however, we focus the discussion on the methods that do not require fully-annotated training data, which is typically expensive to obtain.

In this context, semi-supervised approaches have been investigated. In particular,~\cite{viaflow,supervoxel} proposed to propagate pixel-level annotations provided in the first frame of the sequence throughout the entire video. While this still requires complete annotations in one frame per video,~\cite{viastroke} relied on user scribbles to define foreground and background regions. None of these methods, however, consider background classes. Furthermore, they still all make use of some pixel-level annotations. 

By contrast, weakly-supervised semantic segmentation methods tackle the challenging scenario where only weak annotations, e.g., tags, are given as labels. Much research in this context has been done for still images~\cite{social,tellme,graph-based,multi-image-model,mil,structured,weaklyandsemi,saleh,ccnn,sec,stc,learning,distinct_saliency,fromimagelevel,exploiting}. In particular, most recent methods build on deep networks by making use of objectness criteria~\cite{whatspoint}, object proposals~\cite{fromimagelevel,learning,augmented}, saliency maps~\cite{distinct_saliency,stc,mining,exploiting}, localization cues~\cite{diverse,sec}, convolutional activations~\cite{saleh}, motion cues~\cite{motioncues} and constraints related to the objects~\cite{ccnn,weaklyandsemi}. Since the basic networks have been pre-trained for object recognition, and thus focus on foreground classes, these methods are inherently unable to differentiate multiple background classes.

\begin{figure}[!t]
\centering
\small
\begin{tabular}{c c c}



\scriptsize{Road} & \scriptsize{Tree} &  \scriptsize{Building} \\
\includegraphics[width=.14\textwidth]{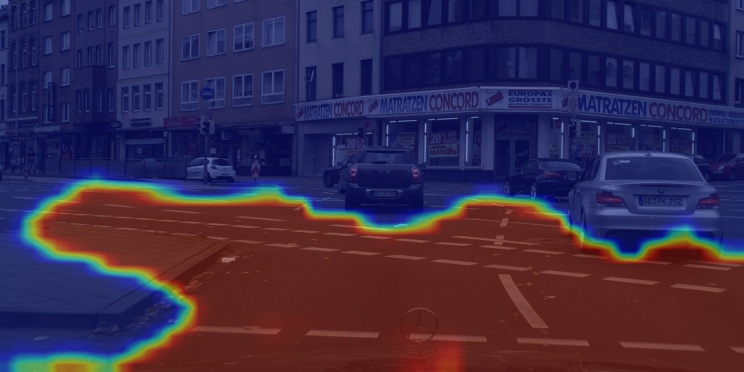} &
\includegraphics[width=.14\textwidth]{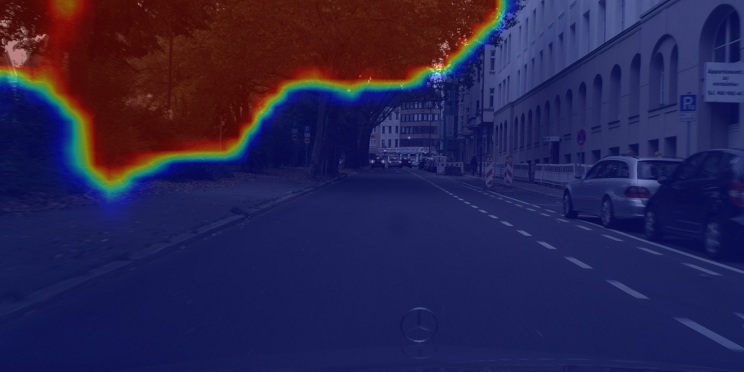} & 
 \includegraphics[width=.14\textwidth]{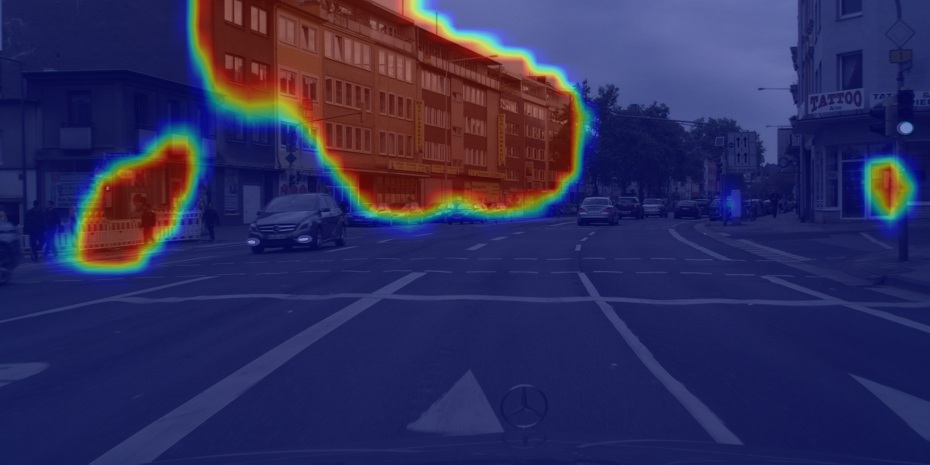} \\

\scriptsize{Bus} & \scriptsize{Bicycle} & \scriptsize{Sign}\\
\includegraphics[width=.14\textwidth]{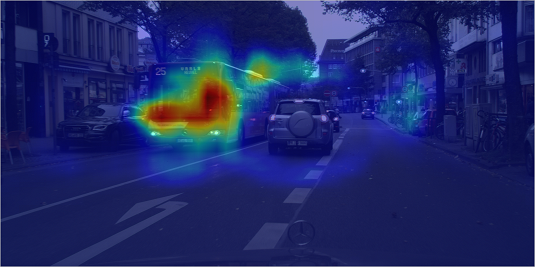} &
\includegraphics[width=.14\textwidth]{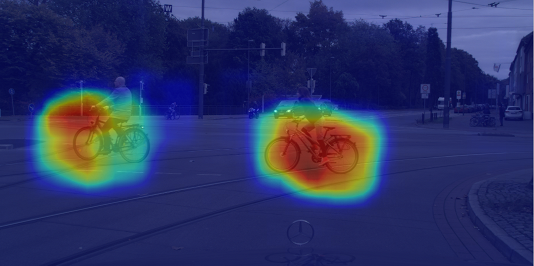} &
\includegraphics[width=.14\textwidth]{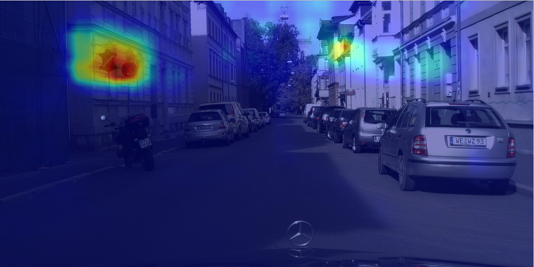} \\

\end{tabular}
\caption{Classifier heatmaps for some of the foreground and background classes of the CityScapes dataset. Note that these heatmaps give a good indication of the location of foreground instances and background regions.} 
\label{fig:heatmap}
\end{figure}

\begin{figure*}[!h]
\centering
\includegraphics[width=0.88\textwidth]{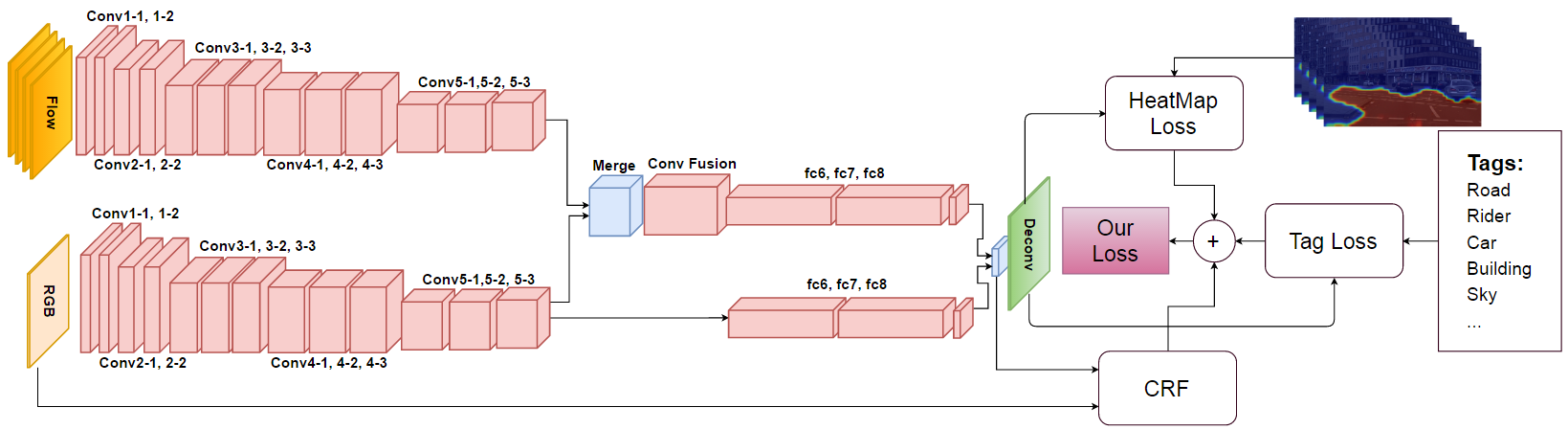}
\caption{\textbf{Proposed Network Structure}. Our two-stream semantic segmentation network leverages both image and optical flow to extract the features. These features are fused in two stages. An early, trainable fusion that puts in correspondence the spatial and temporal information, and a late fusion that merges the resulting spatio-temporal stream with the appearance one for final prediction.}
\label{fig:net}
\end{figure*}

Similarly, most weakly-supervised video semantic segmentation techniques also focus on modeling a single background class. In this context~\cite{webscale,discriminative} work in the even more constrained scenario, where only two classes are considered: foreground vs. background. By contrast, to differentiate multiple foreground classes, but still assuming a single background,~\cite{fast} relied on motion cues and~\cite{webcrawled} made use of a huge amount of web-crawled data (4606 videos with 960,517 frames).

In the same setting of multiple foreground vs. single background, several methods have proposed to rely on additional supervision. For instance,~\cite{viadetection} relied on the CPMC~\cite{cpmc} region detector, which has been trained from pixel-level annotations, to segment foreground from background. In~\cite{domain_adapt} and~\cite{movingobject}, object proposal methods trained from pixel-level and bounding box annotations, respectively, were employed. Similarly,~\cite{obj_track} relied on an object detector trained from bounding boxes. The method of~\cite{co_segment} utilized the FCN trained on PASCAL VOC in a fully-supervised manner to generate initial object segments.

All the weakly-supervised approaches discussed above assume to observe a single background class. In many cases, such as autonomous navigation, however, it is crucial to differentiate the multiple background classes. To the best of our knowledge, only two methods are able to handle this scenario. In~\cite{multi-class}, a nearest-neighbor-based label transfer technique was introduced, which relies on a simple distance between handcrafted features. While this strategy would work well for classes such as grass or sky in which appearance variations are limited, it translates poorly to more challenging and complex scenes, such as urban ones, where individual classes can depict a large range of appearances. As a consequence, this method was only demonstrated on simple scenes containing at most one or two instances of a few classes. In~\cite{co_parsing}, more advanced, deep learning features were exploited. However, this method makes use of  pixel-level supervision to train an FCN to label pixels as either scene vs. object, or multiple scene classes vs. object.

By contrast, we introduce a method that handles multiple foreground and background classes, but only relies on video-level tags. To this end, we introduce a loss function based on classifier heatmaps, and exploit it to train a 
two-stream network jointly leveraging complementary spatial and temporal information in an end-to-end manner.

\section{Our Approach}

In this section, we introduce our approach to weakly-supervised video semantic segmentation. First, we introduce the classifier heatmaps that allow us to model both multiple foreground and background classes. We then introduce our two-stream architecture, which jointly leverages motion and appearance, and discuss our learning scheme, including our loss based on the classifier heatmaps.

\subsection{Classifier Heatmaps}
\label{sec:heatmap}
One of the main challenges when working with tags only for weakly-supervised semantic segmentation is that the annotations do not provide any information about the location of the different classes. While mitigated in the presence of only few foreground classes and a single background one, this problem becomes highly prominent when dealing with complex urban scenes containing many instances of each foreground class and several background classes, such as road, grass, buildings. Existing weakly-supervised methods are dedicated to handle multiple foreground objects, but cannot handle multiple background ones, typically because they inherently rely on object recognition networks, which only tackle foreground classes. To address this, we propose to extract class-specific heatmaps that localize the different classes. Our goal here is to achieve this for both foreground and background classes, and without requiring any pixel-level or bounding box annotations.

Prior work has shown that ConvNets trained with a classification loss can yield remarkable localization results~\cite{oquab2015object,zhou2016learning}. Hence, similarly, for foreground classes, we make use of the VGG-16 network~\cite{vgg} trained on the standard 1000 ImageNet classes. Specifically, we transform the VGG-16 model into a fully-convolutional network by converting its fully-connected layers into convolutional ones, while keeping the trained weights. In other words, the output of the last layer of the transformed model becomes a $W \times H \times 1000$ tensor, and passing an image through the network yields a map showing the activation of each class at each pixel in a low-resolution version of the input image. In practice, we can then access the activations of the foreground classes of interest by only considering a subset of the 1000 ImageNet classes.

The standard 1000 ImageNet classes, however, do not include background. To this end, we collected iconic background images by crawling the background classes on the ImageNet website~\cite{image-net}. We then trained one-vs-all VGG-16 models (pre-trained on the standard 1000 ImageNet classes) for these background classes and followed the same strategy as for the foreground ones to obtain heatmaps. More details are provided in section~\ref{sec:dataset}.

In Fig.~\ref{fig:heatmap}, we show the heatmaps for some of the foreground and background classes of the CityScapes~\cite{Cityscapes} dataset. Note that, while sometimes a bit coarse, the heatmaps still provide valuable information about the location of these classes. In the next section, we introduce our two-stream network that jointly leverages appearance and motion, and show how our heatmaps can be used to train it.

\subsection{Weakly-supervised Two-stream Network}
\label{sec:netStructure}

Videos have two intrinsic features: \emph{Appearance} and \emph{Motion}. To leverage these two sources of information, inspired by the approach of~\cite{2stream} for action recognition, we develop the two-stream network depicted in Fig.~\ref{fig:net}. One stream takes an RGB image as input, and the other optical flow. Compared to taking a series of images as input, explicitly using optical flow to represent motion has the advantage of relieving the network from having to estimate motion implicitly. Below, we discuss how we encode optical flow and describe our fusion strategy. We then introduce our weakly-supervised learning framework.

\vspace{-0.3cm}
\paragraph{Encoding Optical Flow.}
Dense optical flow~\cite{optical-flow} can be represented as a displacement vector field between a pair of frames at time $t$ and $t + 1$. The horizontal and vertical components of the displacement vector field can be thought of as image channels, which makes them well suited to act as input to a convolutional network, such as the one shown in the upper stream of the model in Fig.~\ref{fig:net}. To represent the motion across a video clip, we stack the flow channels corresponding to both directions (vertical and horizontal) of $L$ consecutive frames, in range $(t - \floor{\frac{L}{2}},t + \floor{\frac{L}{2}}]$, to form a total of $2L$ input channels.

\vspace{-0.3cm}
\paragraph{Fusing Appearance and Motion.}
As can be seen in Fig.~\ref{fig:net}, the appearance and motion streams both consist of a series of convolutional layers, following the VGG-16 architecture~\cite{vgg}. The outputs of these streams are then fused at two different levels. In particular, fusion occurs after the fifth convolutional layer (Conv5-3) of each stream, which has been shown to contain a rich semantic representation of the input~\cite{saleh,deepedge}. The first, early fusion puts in correspondence the activations of both streams corresponding to the same pixel location. As~\cite{2stream}, instead of performing sum- or max-fusion, we rely on a convolutional fusion strategy. This gives more flexibility to the network and allows it to learn which channels from the motion and appearance streams should be combined together. The second, late fusion of our network merges the spatio-temporal stream resulting from early fusion with the appearance stream. This fusion is achieved at the point where each stream predicts class scores. The rationale behind this is that the appearance stream provides valuable semantic information on its own, and should thus be propagated to the end of the network. The resulting scores are then passed through a deconvolution layer to obtain the final, full-resolution, semantic map.

\subsubsection{Weakly-Supervised Learning}
We now introduce our learning algorithm for weakly-supervised semantic segmentation. We first introduce a simple loss based on image tags only, and then show how we can incorporate the localization information of our classifier heatmaps to the loss function.

Intuitively, given image tags, one would like to encourage the image pixels to be labeled as one of the classes that are observed in the image, while preventing them to be assigned to unobserved classes. Note that this assumes that the full set of tags available cover all the classes depicted in the image, which is a common assumption in weakly-supervised semantic segmentation~\cite{ccnn,weaklyandsemi,mil,whatspoint,sec,saleh}.
Formally, given an input video $V$, let $\mathcal{L}$ be the set of tags associated to $V$ and $\bar{\mathcal{L}}$ the class labels that are not among the tags. Furthermore, let us denote by $s_{i,j}^k(\theta)$ the score produced by our network with parameters $\theta$ for the pixel at location ($i,j$) and for class $k$, $0\leq k < N$, in the current input video frame $I$. Note that, in general, we will omit the explicit dependency of the variables on the network parameters. Finally, let $S_{i,j}^k$ be the probability of class $k$ obtained after a softmax layer, i.e., 
\begin{equation}
S_{i,j}^k = \frac{\exp(s_{i,j}^k)}{\sum_{c=1}^N\exp(s_{i,j}^c)}\;.
\end{equation}

Encoding the above-mentioned intuition can then simply be achieved by designing a loss of the form
\begin{equation}
L_{tag} = -\frac{1}{|\mathcal{L}|}\sum_{k\in \mathcal{L}} \log{S^k} - \frac{1}{|\bar{\mathcal{L}}|}\sum_{k\in \bar{\mathcal{L}}}\log(1-S^k) \;,
\label{eq:loss_mil}
\end{equation}
where $S^k$ represents a candidate score for each class in the input frame. In short, the first term in Eq.~\ref{eq:loss_mil} expresses the fact that the present classes should be in the input frame, while the second term penalizes the pixels that have high probabilities for the absent classes. In practice, instead of computing $S^k$ as the maximum probability (as previously used in~\cite{mil,whatspoint}) for class $k$ over all pixels in the input frame, we make use of the convex Log-Sum-Exp (LSE) approximation of the maximum (as previously used in~\cite{fromimagelevel,saleh}), which can be written as
\begin{equation}
\tilde{S}^k = \frac{1}{r}\log\left[\frac{1}{|I|}\sum_{i,j \in I}\exp(rS_{i,j}^k)\right]\;,
\label{eq:LSE}
\end{equation}
where $|I|$ denotes the total number of pixels in the input frame and $r$ is a parameter allowing this function to behave in a range between the maximum and the average. In practice, following~\cite{fromimagelevel,saleh}, we set $r$ to 5.

The loss of Eq.~\ref{eq:loss_mil} does not rely on any localization cues. As a consequence, minimizing it will typically yield poor object localization accuracy. To overcome this issue, we propose to make use of the classifier heatmaps introduced in Section~\ref{sec:heatmap}. To this end, we first generate binary masks $B_k$ for each class $k$. These binary masks are obtained by setting to 1 the values that are above 20\% of the maximum value in the heatmap of class $k$, and to 0 the other ones. 

Our goal then is to encourage the model to have, for each class, high probability at pixels inside the corresponding binary mask. To this end, we introduce the loss function 
\begin{equation}
L_{heatmap} = -\frac{1}{|\mathcal{L}|}\sum_{k\in \mathcal{L}}\frac{1}{|{B_k}|}\sum_{{i,j}\in{B_k}} \log{S_{i,j}^k}\;,
\label{eq:loss_heatmap}
\end{equation}
which we use in conjunction with the loss of Eq.~\ref{eq:loss_mil}.

While this heatmap-based loss significantly helps localizing the different classes, the heatmaps typically only roughly match the class boundaries. To overcome this, we follow the CRF-based strategy of~\cite{sec}. Specifically, we construct a fully-connected CRF, with
unary potentials corresponding to the probability scores predicted by our segmentation network, and image-dependent Gaussian pairwise potentials~\cite{dcrf}. 
We then add another term to the loss function, corresponding to the mean KL-divergence between the outputs of the network and the outputs of the fully connected CRF. This term encourages the network prediction to coincide with the CRF output, which produces segmentations that better respect the image boundaries.

Altogether, our network can handle multiple foreground and background classes, and, as discussed in more detail in Section~\ref{sec:details}, can be trained in an end-to-end fashion.

\section{Experiments}
In this section, we first describe the datasets used in our experiments and provide details about our learning and inference procedures. We then present the results of our model and compare it to state-of-the-art weakly-supervised semantic segmentation methods.

\subsection{Datasets}
\label{sec:dataset}
To demonstrate the effectiveness of our approach, and evaluate the different components of our model, we use the challenging  CityScapes~\cite{Cityscapes} and CamVid~\cite{camvid} road scene datasets. Furthermore, to compare to the state-of-the-art, we make use of YouTube-Objects~\cite{youtube}, which most weakly-supervised video semantic segmentation methods report on. Note that, although different annotation types are provided in each of these datasets, we only make use of tags, indicating which classes are present in each video clip.

\begin{table}[!t]
\centering
\small
\caption{Background classes used to train our classifiers (Section~\ref{sec:heatmap}) for the CityScapes and CamVid datasets.}
\label{tab:background_samples}
\begin{tabular}{| l|c c c c c c|} 
\hline
Class & \rotatebox[origin=c]{60}{\scriptsize{road}} & \rotatebox[origin=c]{60}{\scriptsize{sidewalk}} & \rotatebox[origin=c]{60}{\scriptsize{building}} & \rotatebox[origin=c]{60}{\scriptsize{vegetation}} & \rotatebox[origin=c]{60}{\scriptsize{terrain}} & \rotatebox[origin=c]{60}{\scriptsize{sky}}\\
 \hline

\#Samples & 126 & 306 & 670 & 176 & 190 & 180\\
\hline
\end{tabular}
\vspace{-0.2cm}
\end{table}

\textbf{CityScapes}: Cityscapes~\cite{Cityscapes} is a recently released large-scale dataset, containing high quality pixel-level annotations for 5000 images collected in street scenes from 50 different cities. The images of Cityscapes have resolution 2048$\times$1024, making it a challenge to train very deep networks with limited GPU memory. We therefore downsampled the images by a factor of 2.
The annotations correspond to the 20$^{th}$ frame of 30-frame video snippets. 
We then extracted optical flow from 10 consecutive frames, from the 16$^{th}$ to the 25$^{th}$, and used the RGB frames and image-level tags in conjunction with these optical flows to train our model. 
We made use of the standard training/validation/test partitions, containing 2975,500, and 1525 images, respectively. Following the standard evaluation protocol~\cite{Cityscapes}, we used 19 semantic labels (belonging to 7 super categories: ground, construction, object, nature, sky, human, and vehicle) for evaluation (the void label is not considered for evaluation). 

\textbf{CamVid}: CamVid dataset consists of over 10 minutes of high quality 30 Hz footage. The videos are captured at $960\times 720$ resolution with a camera mounted inside a car. Three of the four sequences were shot in daylight, and the fourth one was captured at dusk. This dataset contains 32 categories. In our experiments, following~\cite{camvid11,featurespace,segnet}, we used a subset of 11 classes. The dataset is split into 367 training, 101 validation and 233 test images. As for CityScapes, ground-truth labels are provided every 30 frames. We extracted optical flow in 10 frames around the labeled ones, and used them with the RGB frames for training. 


\textbf{Iconic Data}: 
The background classes and number of samples per class, extracted from the background images of the ImageNet website, as mentioned in Section~\ref{sec:heatmap}, and used to train our background classifiers for CityScapes and CamVid are given in Table~\ref{tab:background_samples}. Note that, in the standard 1000 classes of ImageNet, there is no general \emph{person} class, which appears in both datasets. To handle this class, we therefore proceeded in a similar manner as for the background classes, but making use of a small subset of the samples (1300 samples) from~\cite{inria,nicta-ped}.

\textbf{YouTube-Objects}: The YouTube-Objects dataset is composed of videos collected from YouTube by querying for the names of 10 object classes of the PASCAL VOC Challenge. It contains between 9 and 24 videos per class. The duration of each video ranges from 30 seconds to 3 minutes. The videos are weakly annotated, with each video containing at least one object of the corresponding queried class. In the dataset, the videos are separated into shots.
For our experiments, we randomly extracted 6-8 frames from each shot to obtain a total of 13800 frames out of 700,000 ones available in the dataset.
We again made use of snippets of 10 frames to encode optical flow. 

For evaluation, we used the subset of images with pixel-level annotations provided by~\cite{mask-yto}. Note that there is no overlap between this subset and the shots from which we extracted the training data.

\subsection{Implementation Details}
\label{sec:details}
To train our two-stream network, introduced in Section~\ref{sec:netStructure}, we relied on stochastic gradient descent with a learning rate starting at $10^{-5}$ with a decrease factor of $10$ 
every $10k$ iterations, a momentum of $0.9$, a weight decay of $0.0005$, and mini-batches of size $1$. Similarly to recent weakly-supervised segmentation methods~\cite{saleh,whatspoint,mil,weaklyandsemi,sec}, the weights of our two-stream network were initialized with those of the 16-layer VGG classifier~\cite{vgg} pre-trained for 1000-way classification on the ILSVRC 2012~\cite{ILSVRC}. Hence, for the last convolutional layer, we used the weights corresponding to the classes shared by the datasets used here and in ILSVRC. For the background classes, we initialized the weights with zero-mean Gaussian noise with a standard deviation of $0.1$.
At inference time, given only the test image and optical flow, the network generates a dense prediction as a complete semantic segmentation map. 

For both CamVid and CityScapes, we used the GPU implementation of~\cite{optical-flow} to generate the stack of optical flow for each snippet of length 10. For YouTube-Objects, we used the optical flow information pre-computed by~\cite{flow-yto}. Note that neither of these methods relies on any learning strategy, and thus they can be directly applied to our input images.
We used C++ and Python (the Caffe framework~\cite{caffe}) for our implementation. As other methods~\cite{sec,weaklyandsemi,saleh,ccnn,stc}, we further applied a dense CRF~\cite{dcrf} to refine this initial segmentation. 
To this end, we used the default CRF parameter values as in the original paper~\cite{dcrf}.

\subsection{Experimental Results}

Below, we first evaluate the different components of our method on the validation set of the two challenging road scene datasets. We then provide results of our complete framework on their respective test sets. Finally, we compare our approach to state-of-the-art weakly-supervised segmentation methods on YouTube-Objects.

\vspace{-0.3cm}
\subsubsection{Ablation Study}
To evaluate the influence of the different components of our approach, we designed the following baselines. \emph{No-Heatmap} corresponds to a single-stream model exploiting the RGB image only, without exploiting our heatmap-based loss of Eq.~\ref{eq:loss_heatmap},
i.e., using the loss of Eq.~\ref{eq:loss_mil} and the CRF loss. \emph{Foreground-Heatmap} consists of a similar single stream network, additionally using the loss of Eq.~\ref{eq:loss_heatmap}, but only for the foreground classes extracted from the VGG-16 network pre-trained on ILSVRC. \emph{Our-Heatmap} corresponds to using all our heatmaps, i.e., for foreground and background classes, with a single-stream network. Finally, \emph{Ours} corresponds to our two-stream network with all the loss terms. 

We report the results of these different models in Table~\ref{tab:cityscapes_setups} for Cityscapes and in Table~\ref{tab:camvid_setups} for CamVid. In particular, we report the mean Intersection over Union (mIoU), the average per-class accuracy and global accuracy. The general behavior is the same for both datasets: Exploiting heatmaps for foreground class improves over not using heatmaps at all. However, also relying on heatmaps for background classes gives a significant boost in performance. Finally, jointly leveraging appearance and motion in our two-stream network further improves segmentation accuracy. As can be observed in Table~\ref{tab:eval_heatmap_camvid}, which provides the per-class intersection over union for CamVid, our heatmaps and our two-stream network add significant improvement to the baselines for most of the classes, especially in background classes, e.g., sky and road. 

\begin{table}[!t]
\renewcommand{\arraystretch}{1.4}
\centering
\scriptsize
\caption{Influence of our heatmaps and of optical flow. These results were obtained using the CityScapes validation set.}
\label{tab:cityscapes_setups}

\begin{tabular}{l| c | c | c }
\hline
 Setup & Mean IOU & Mean Class Acc. & Global Acc. \\ 
 \hline
 No-Heatmap & \small{8.4\%} & \small{18.8\%} & \small{20.9\%}  \\
 ImageNet-Heatmap & \small{11.4\%} & \small{33.2\%} & \small{22.0\%} \\
 Our-Heatmap & \small{20.6\%} & \small{40.6\%} & \small{54.0\%} \\
 Our Two-Stream & \small{23.6\%} & \small{40.3\%} & \small{63.9\%} \\
 \hline
\end{tabular}
\end{table}

\begin{table}[!t]
\renewcommand{\arraystretch}{1.4}
\centering
\scriptsize
\caption{Influence of our heatmaps and of optical flow. These results were obtained using the CamVid validation set.}
\label{tab:camvid_setups}

\begin{tabular}{l| c | c | c }
\hline
 Setup & Mean IOU & Mean Class Acc. & Global Acc.\\ 
 \hline
 No-Heatmap & \small{10.2\%} & \small{24.9\%} & \small{19.5\%} \\
 ImageNet-Heatmap & \small{11.0\%} & \small{25.8\%} & \small{28.9\%} \\
 Our-Heatmap & \small{29.5\%} & \small{49.7\%} & \small{62.6\%} \\
 Our Two-Stream &  \small{31.1\%} & \small{50.2\%} & \small{67.4\%} \\
 \hline
\end{tabular}
\end{table}

\begin{table*}[!t]
\renewcommand{\arraystretch}{1.2}
\centering
\small
\caption{Influence of our heatmaps and of optical flow. Per-class IoU for the CamVid validation set.}
\label{tab:eval_heatmap_camvid}
\scalebox{0.85}
{
\hspace{-0.2cm}
\begin{tabular}{l | c c c c c c c c c c c} 
\hline
Setup & \rotatebox[origin=c]{90}{\scriptsize{building}} & \rotatebox[origin=c]{90}{\scriptsize{vegetation}} & \rotatebox[origin=c]{90}{\scriptsize{sky}} & \rotatebox[origin=c]{90}{\scriptsize{car}} & \rotatebox[origin=c]{90}{\scriptsize{sign}} & \rotatebox[origin=c]{90}{\scriptsize{road}} & \rotatebox[origin=c]{90}{\scriptsize{pedestrian}} & \rotatebox[origin=c]{90}{\scriptsize{fence}} & \rotatebox[origin=c]{90}{\scriptsize{pole}} & \rotatebox[origin=c]{90}{\scriptsize{sidewalk}} & \rotatebox[origin=c]{90}{\scriptsize{cyclist}}\\
 \hline
 No-HeatMap & 37.0 &   33.0 &   0.0 &  28.6 &   7.8 &    4.6 &   0.0 &   0.1 &   0.7 &   0.4 &   0\\
ImageNet-HeatMap & 29.8 &  0.0 &    0.1 &   14.1 &    7.5 &   53.4 &    4.9 &    4.9 &    0.2 &    0.0 &    6.2\\
Our-HeatMap & 54.1 &   76.1 &   86.3 &   19.4 &    6.6 &   56.3 &    9.0 &    0.9 &    0.5 &    6.0 &    9.0\\
 Our Two-Stream & 63.4 &   72.2 &   84.2 &   19.3 &    8.9 &   60.6 &   14.3 &    0.0 &    0.0 &    4.1 &   15.2\\
\hline

 \end{tabular}
 }
 \end{table*}

\begin{table*}[!tp]
\renewcommand{\arraystretch}{1.2}
\centering
\small
\caption{Comparison to fully-supervised semantic segmentation methods on the CamVid test set. While we use the weakest level of supervision, the difference to fully supervised methods, especially in background classes (sky, building, road and tree) is remarkably low.
}
\label{tab:eval_detail_camvid}
\scalebox{0.9}
{
\hspace{-0.2cm}
\begin{tabular}{l | l | c c c c c c c c c c c  | c } 
\hline
 Method & Supervision & \rotatebox[origin=c]{90}{\scriptsize{building}} & \rotatebox[origin=c]{90}{\scriptsize{vegetation}} & \rotatebox[origin=c]{90}{\scriptsize{sky}} & \rotatebox[origin=c]{90}{\scriptsize{car}} & \rotatebox[origin=c]{90}{\scriptsize{sign}} & \rotatebox[origin=c]{90}{\scriptsize{road}} & \rotatebox[origin=c]{90}{\scriptsize{pedestrian}} & \rotatebox[origin=c]{90}{\scriptsize{fence}} & \rotatebox[origin=c]{90}{\scriptsize{pole}} & \rotatebox[origin=c]{90}{\scriptsize{sidewalk}} & \rotatebox[origin=c]{90}{\scriptsize{cyclist}} &  \rotatebox[origin=c]{0}{mIOU}\\
 \hline
SegNet~\cite{segnet} & pixel level annotation & 68.7 & 52.0 & 87.0 & 58.5 & 13.4 & 86.2 & 25.3 & 17.9 & 16.0 & 60.5 & 24.8  & 46.4\\
Liu and He~\cite{activeinference} & pixel level annotation& 66.8 & 66.6 & 90.1 & 62.9 & 21.4 & 85.8 & 28.0 &17.8 & 8.3 & 63.5 & 8.5 & 47.2 \\
\hline
Ours & image-level tags & 58.9 &   46.4 & 83.8 &   26.5 &   12.0 &   64.4 &    8.0 &   11.3 & 3.1 &   1.1  & 11.0 & 29.7\\
 \hline
 \end{tabular}
 }
 \end{table*}

\begin{table*}[!t]
\renewcommand{\arraystretch}{1.3}
\centering
\scriptsize
\caption{Comparison to fully-supervised semantic segmentation methods on the CityScapes test set. As on CamVid, while we use the weakest level of supervision, the gap with fully supervised methods is quite low, particularly on background classes.
}
\label{tab:eval_detail_cityscapes}
\scalebox{0.8}
{
\hspace{-0.2cm}
\begin{tabular}{l| l | c c c c c c c c c c c c c c c c c c c | c | c } 
\hline
Method & Supervision & \rotatebox[origin=c]{90}{road} & \rotatebox[origin=c]{90}{sidewalk} & \rotatebox[origin=c]{90}{building} & \rotatebox[origin=c]{90}{wall} & \rotatebox[origin=c]{90}{fence} & \rotatebox[origin=c]{90}{pole} & \rotatebox[origin=c]{90}{traffic-light} & \rotatebox[origin=c]{90}{traffic-sign} & \rotatebox[origin=c]{90}{vegetation} & \rotatebox[origin=c]{90}{terrain} & \rotatebox[origin=c]{90}{sky} & \rotatebox[origin=c]{90}{person} & \rotatebox[origin=c]{90}{rider} & \rotatebox[origin=c]{90}{car} & \rotatebox[origin=c]{90}{truck} & \rotatebox[origin=c]{90}{bus} & \rotatebox[origin=c]{90}{train} & \rotatebox[origin=c]{90}{motorcycle}& \rotatebox[origin=c]{90}{bicycle} & \rotatebox[origin=c]{90}{mIOU Class}& \rotatebox[origin=c]{90}{mIOU Category}\\
 \hline
 
 FCN-8s~\cite{FCN}& pixel-level annotation & 97.4 & 78.4 & 89.2 & 34.9 & 44.2 & 47.4 & 60.1 & 65.0 & 91.4 & 69.3 & 93.9 & 77.1 & 51.4 & 92.6 & 35.3 & 48.6 & 46.5 & 51.6 & 66.8 & 65.3 & 85.7\\
 Deeplab~\cite{deeplab}& pixel-level annotation &97.3 &	77.6 &	87.7 &	43.6 &40.4 &	29.7 &	44.5 &	55.4 &	89.4 &	67.0 &	92.7 &	71.2 &	49.4 &	91.4 &	48.7 &	56.7 &	49.1 &	47.9 & 58.6 & 63.1 & 81.2\\
 SegNet~\cite{segnet}&pixel-level annotation & 96.4	& 73.2 &	84.0 &	28.4 &	29.0 &	35.7 &	39.8 &	45.1 &	87.0 &	63.8 &	91.8 & 	62.8 &	42.8 &	89.3 &	38.1 &	43.1 &	44.1 &	35.8 &	51.9 & 56.9 & 79.1 \\
 \hline
 Ours& image-level tags &78.5 &	2.7 &	45.0 &	6.6 &	9.8 &	5.4 &	0.7 &	2.1 &	63.3 &	22.0 &	71.5 &	17.6 &	8.0 &	43.6 &	16.0 &	15.5 &	33.0 &	17.9 &	13.6 & 24.9 & 47.2\\
 \hline
 \end{tabular}
 }
 \end{table*}
 
\begin{table*}[h!]
\renewcommand{\arraystretch}{1.2}
\small
\centering
\caption{Comparison to the state-of-the-art on the YouTube-Objects dataset. We report the per-class and mean IoU. Note that our two-stream network significantly outperforms the state-of-the-art baselines.}
\label{tab:yto}
\scalebox{0.99}
{
\hspace{-0.2cm}\begin{tabular}{ l|c c c c c c c c c c |c } 
\hline
Method & \rotatebox[origin=c]{0}{aeroplane} & \rotatebox[origin=c]{0}{bird} & \rotatebox[origin=c]{0}{boat} & \rotatebox[origin=c]{0}{car} & \rotatebox[origin=c]{0}{cat} & \rotatebox[origin=c]{0}{cow} & \rotatebox[origin=c]{0}{dog} & \rotatebox[origin=c]{0}{horse} & \rotatebox[origin=c]{0}{motorbike} & \rotatebox[origin=c]{0}{train} &  \rotatebox[origin=c]{0}{mIOU}\\
 \hline
Papazoglou et al.~\cite{fast} & 67.4 & 62.5 & 37.8  & 67.0 & 43.5 & 32.7 & 48.9 & 31.3 & 33.1 & 43.4& 46.8\\
Tang et al.~\cite{discriminative} & 17.8 & 19.8 & 22.5 & 38.3 & 23.6 & 26.8 & 23.7 & 14.0& 12.5 & 40.4 & 23.9\\
Ochs et al.~\cite{long-term-video}&13.7 & 12.2 & 10.8 & 23.7 & 18.6 & 16.3 & 18.0 & 11.5 & 10.6&  19.6 & 15.5\\
\hline

 Ours &   67.6 &   72.3 &   58.1 &   60.1 &   59.8 &   42.6 &   60.1 &   46.3 &  53.6 &   12.4 & 53.3 \\

\hline
\end{tabular}
}
\end{table*}

Furthermore, we evaluated the influence of the CRF on our results. On CityScapes, our two-stream network without the CRF loss achieves 20.3\% mIOU vs 23.6\% with the CRF, thus showing that the CRF helps, but is not the key to our results.

Regarding runtimes, the average inference time of our method per image on CityScapes given optical flow is 0.56s without CRF inference as post-processing and 3.6s with CRF inference. This matches the runtimes reported in other papers that worked on CityScapes, although in the fully-supervised setting, such as~\cite{FCN} (0.5s without CRF) and~\cite{deeplab} (4s with CRF).

\subsubsection{Results on Test Sets}
We then evaluated our complete approach on the test sets of CamVid and CityScapes. In Table~\ref{tab:eval_detail_camvid} and Table~\ref{tab:eval_detail_cityscapes}, we compare the results of our weakly-supervised approach to those of fully-supervised methods. Note that, while these methods make use of much stronger supervision during training, thus making the comparison unfair to us, the gap in accuracy with our method, especially for background classes (sky, building, road and tree) is remarkably low. This further illustrates the strength of our approach, which, despite using only tags, yields good segmentation accuracy.

Qualitative results of our two-stream network on samples from CityScapes and CamVid are also depicted in Fig.~\ref{fig:result}.

\subsubsection{Comparison to the State-of-the-Art}
\begin{figure*}[p!]
\centering
\small
\begin{tabular}{c c c c }
\multicolumn{4}{c}{CityScapes Dataset}\\
\includegraphics[width=.21\textwidth]{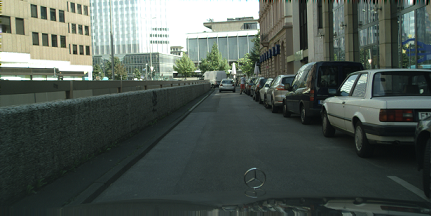}&
\includegraphics[width=.21\textwidth]{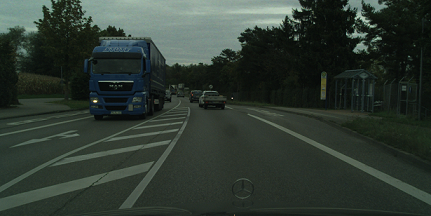}&
\includegraphics[width=.21\textwidth]{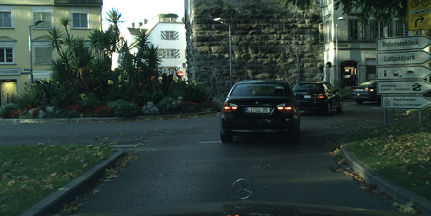}&
\includegraphics[width=.21\textwidth]{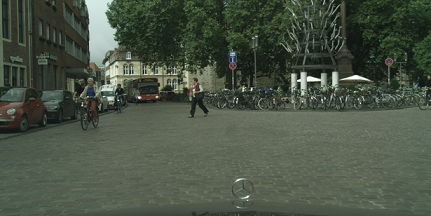}\\
\includegraphics[width=.21\textwidth]{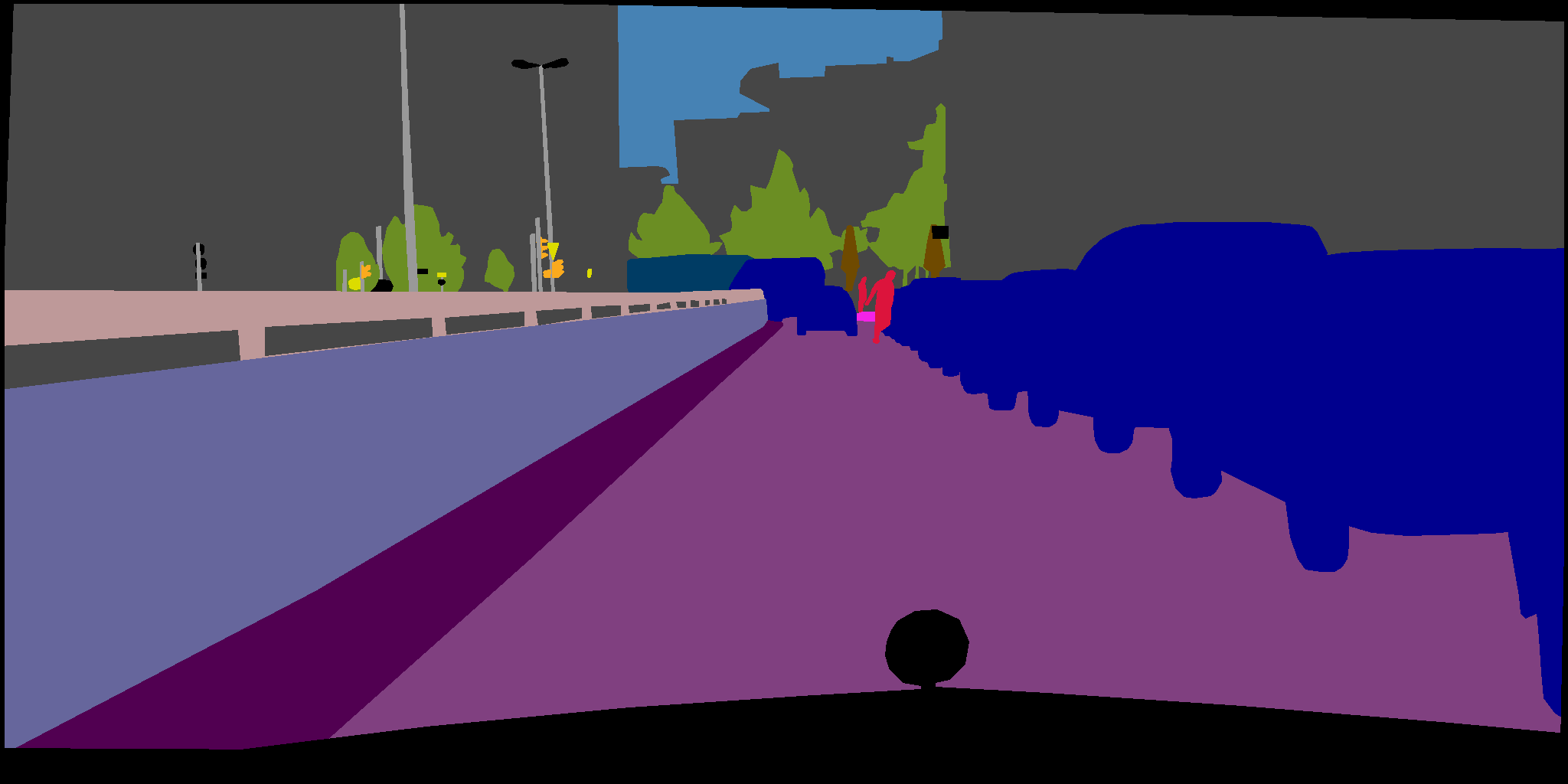}&
\includegraphics[width=.21\textwidth]{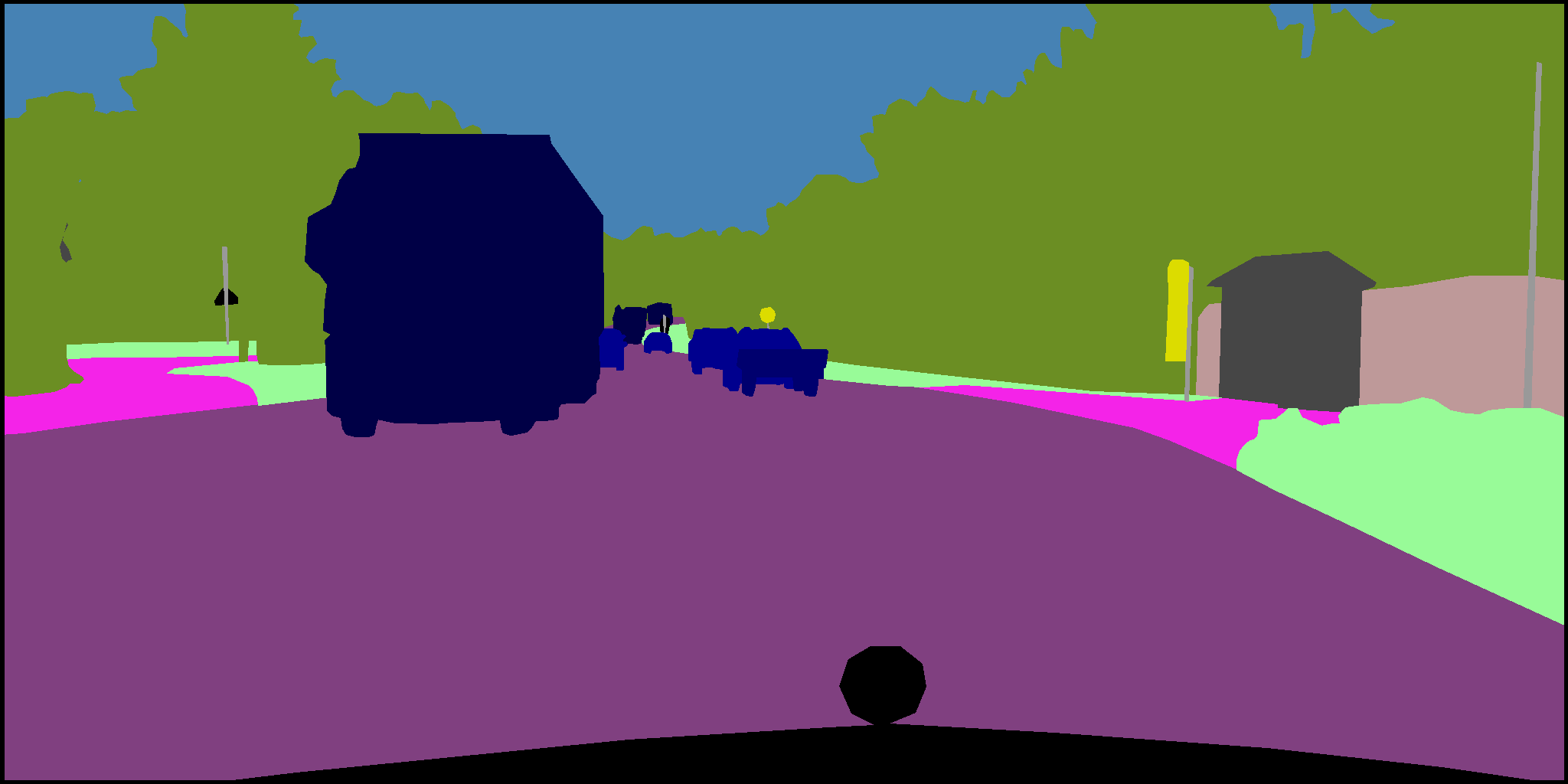}&
\includegraphics[width=.21\textwidth]{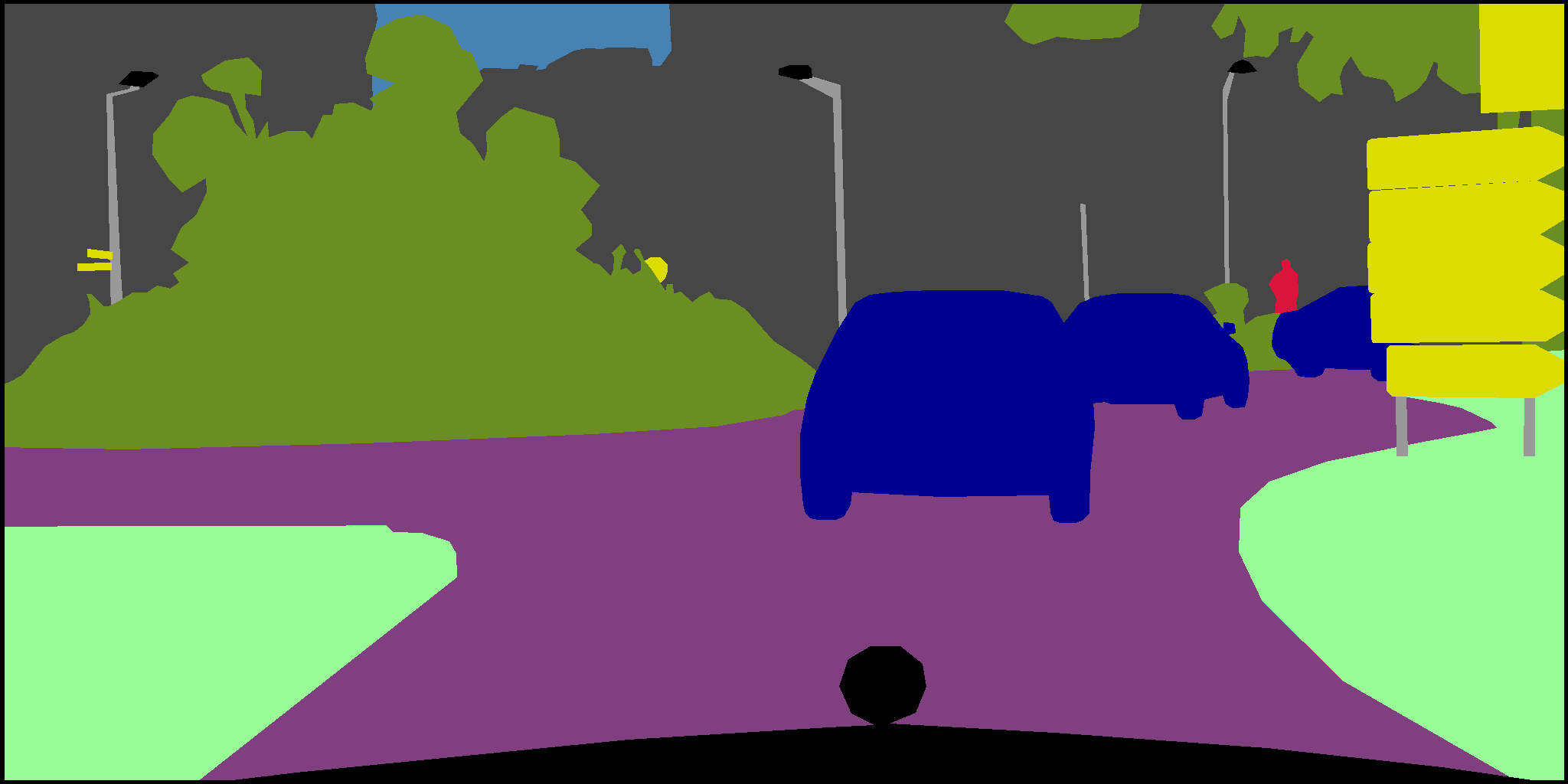}&
\includegraphics[width=.21\textwidth]{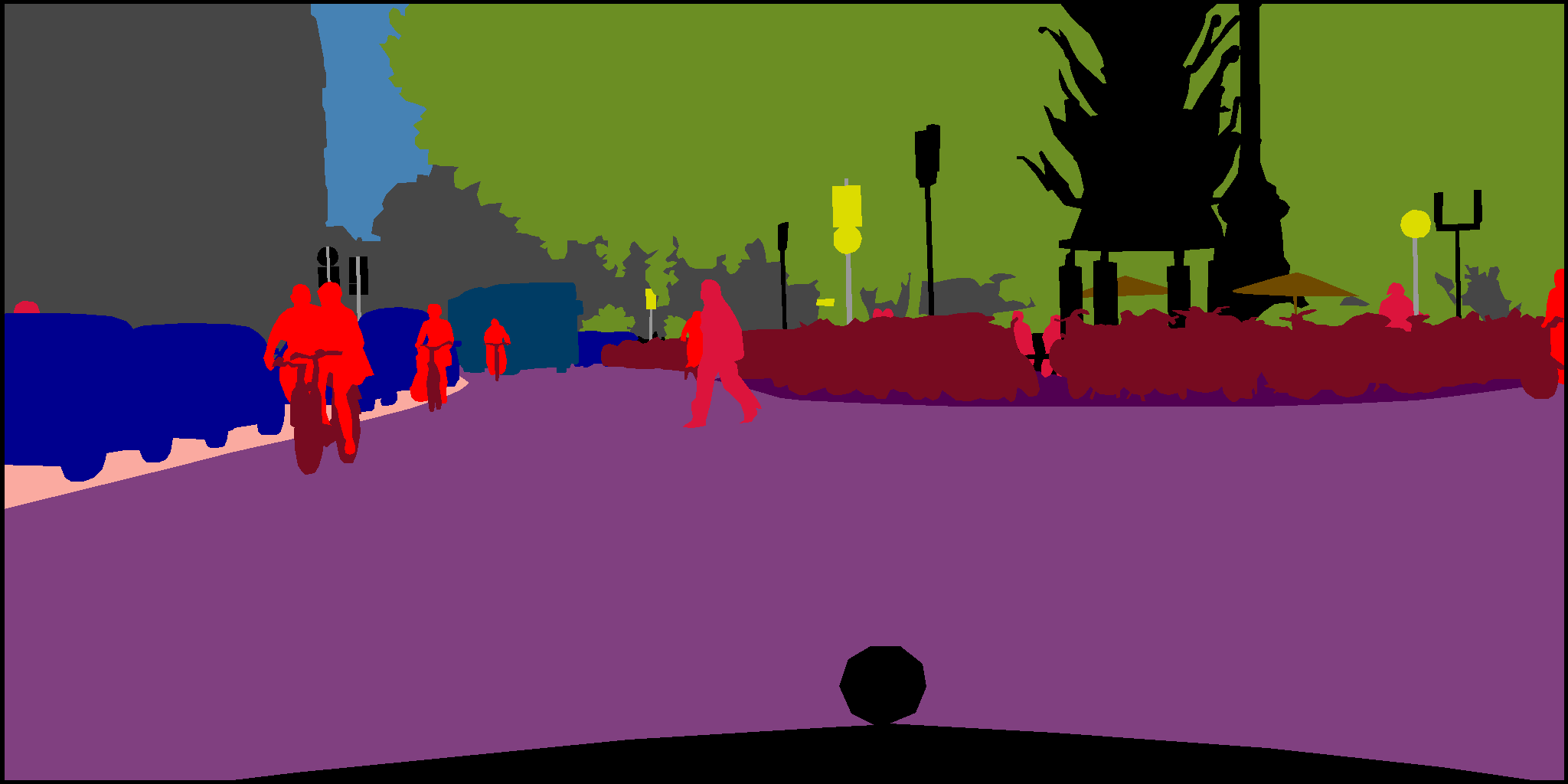}\\
\includegraphics[width=.21\textwidth]{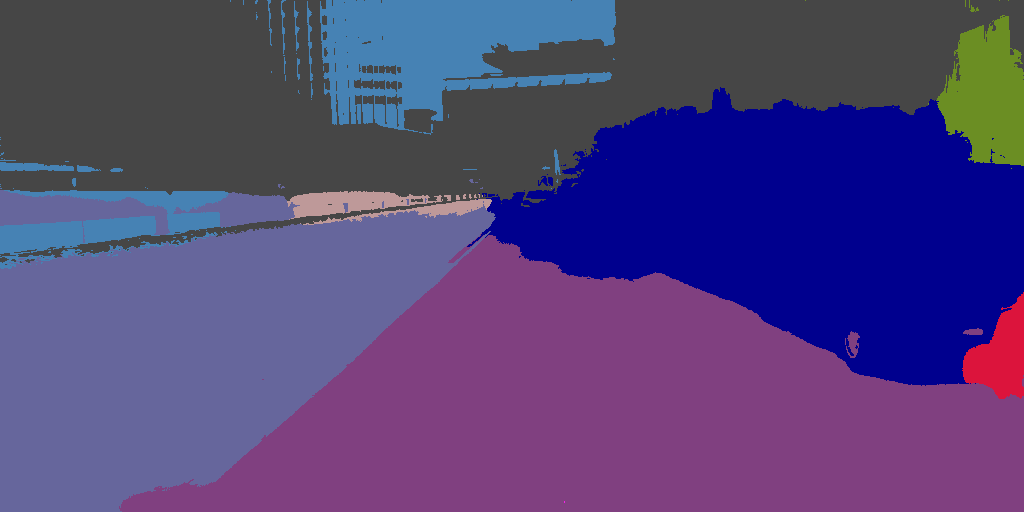}&
\includegraphics[width=.21\textwidth]{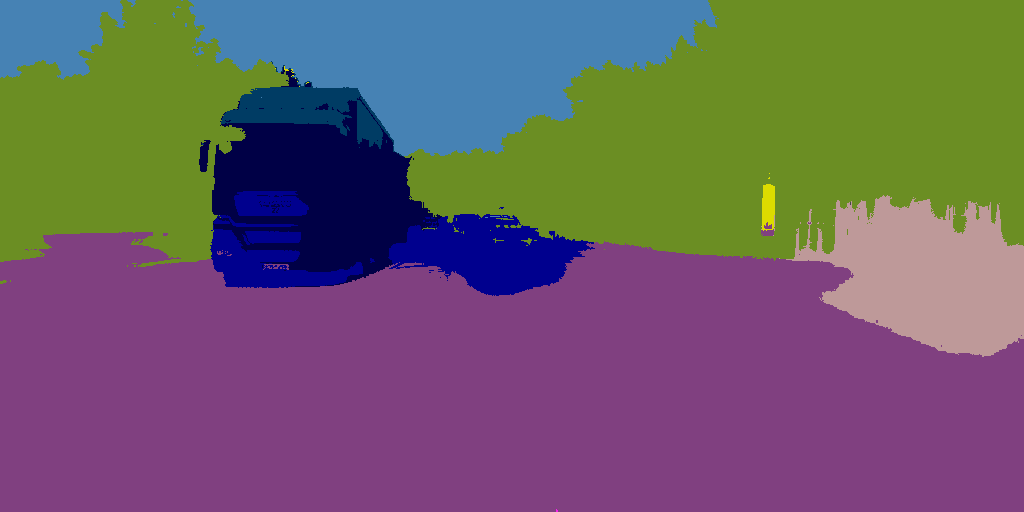}&
\includegraphics[width=.21\textwidth]{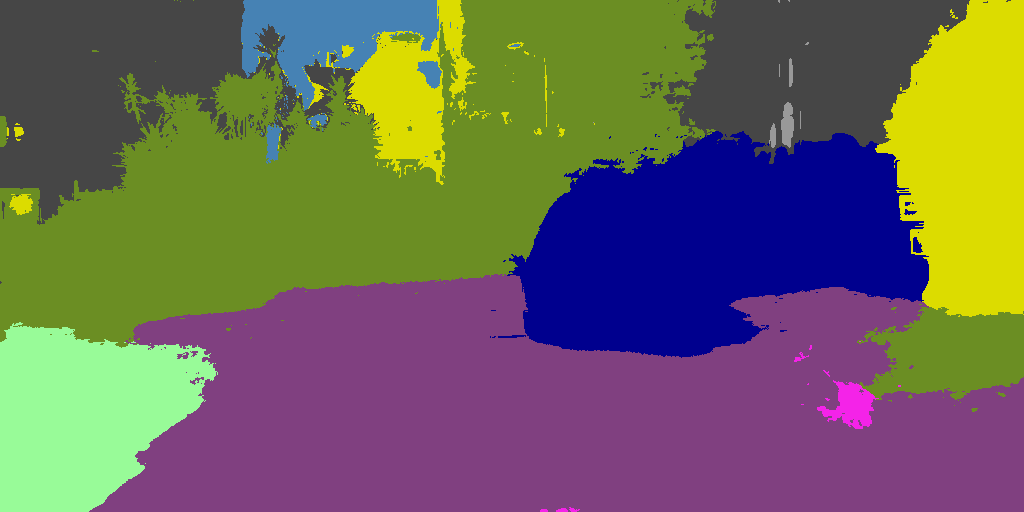}&
\includegraphics[width=.21\textwidth]{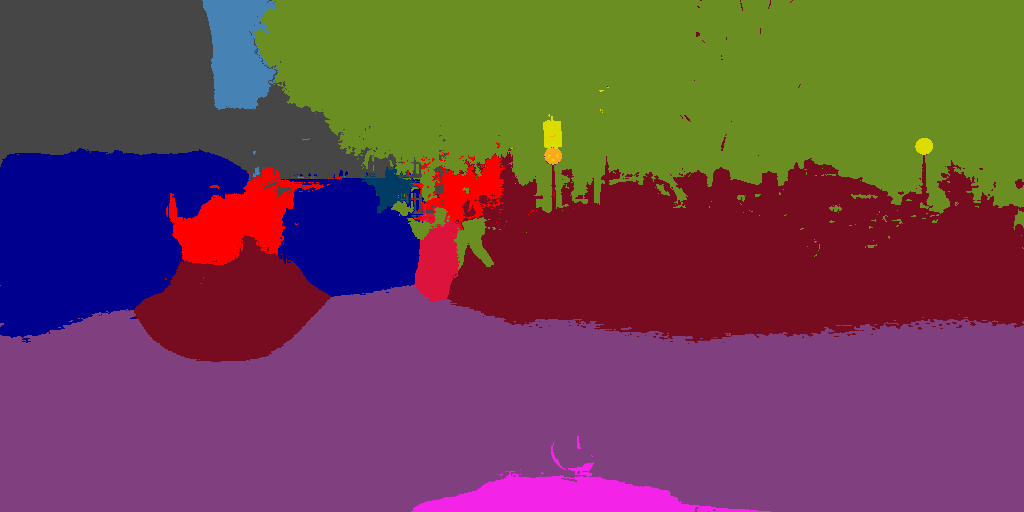}\\
\multicolumn{4}{c}{CamVid Dataset}\\
\includegraphics[width=.21\textwidth]{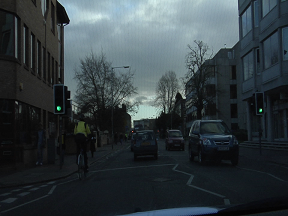}&
\includegraphics[width=.21\textwidth]{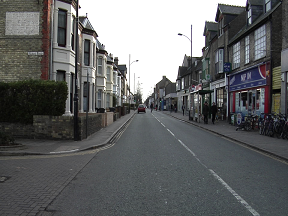}&
\includegraphics[width=.21\textwidth]{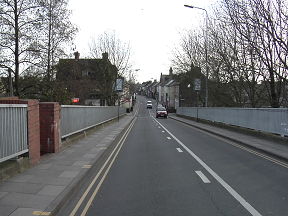}&
\includegraphics[width=.21\textwidth]{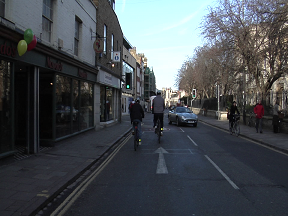}\\
\includegraphics[width=.21\textwidth]{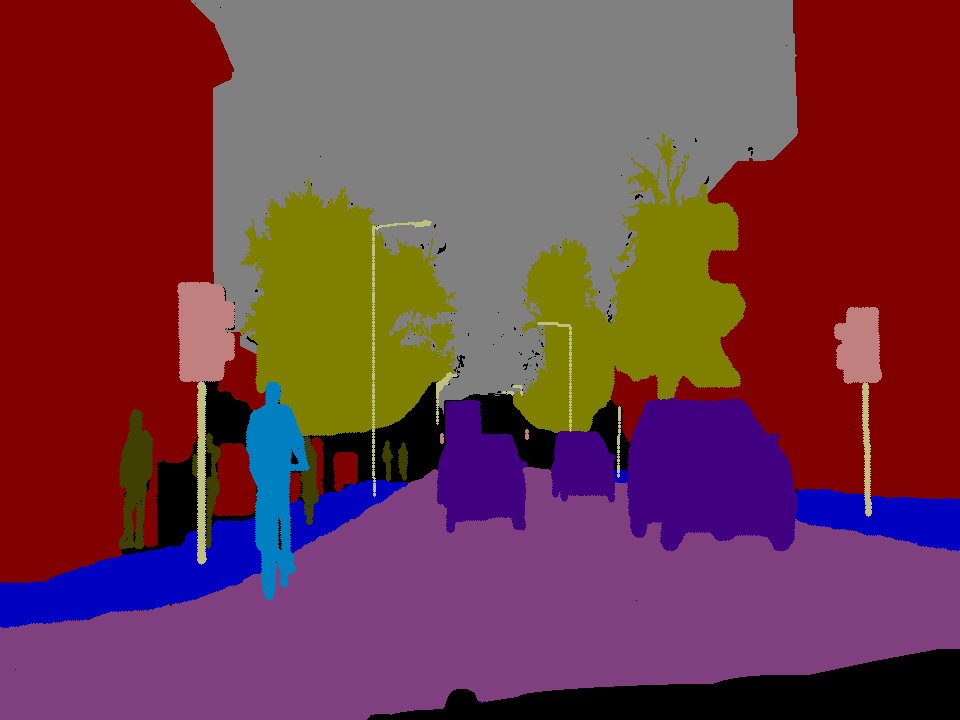}&
\includegraphics[width=.21\textwidth]{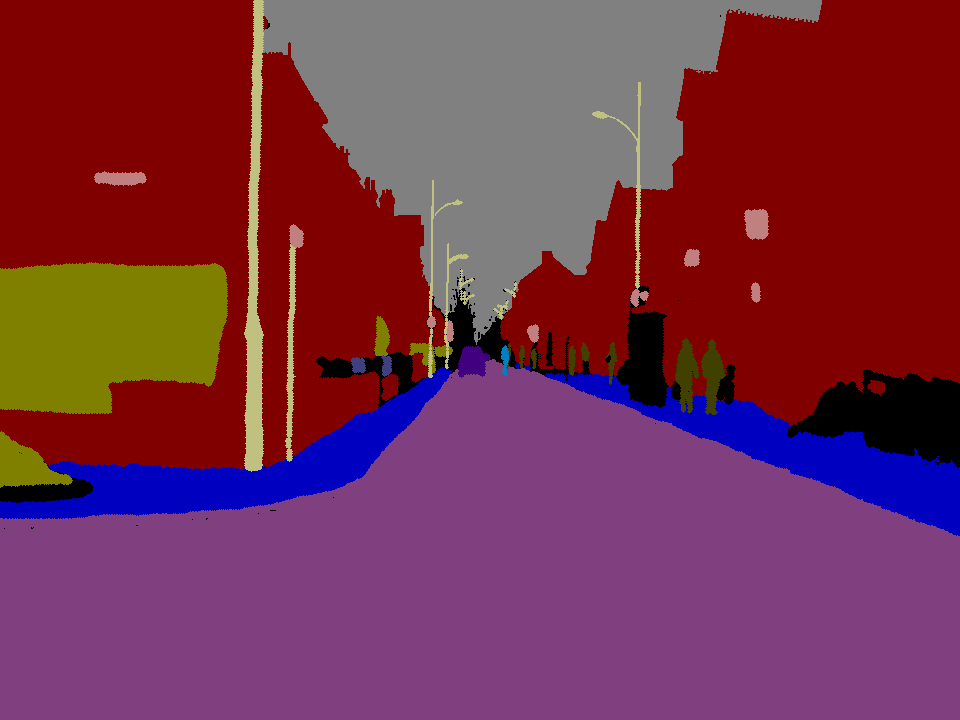}&
\includegraphics[width=.21\textwidth]{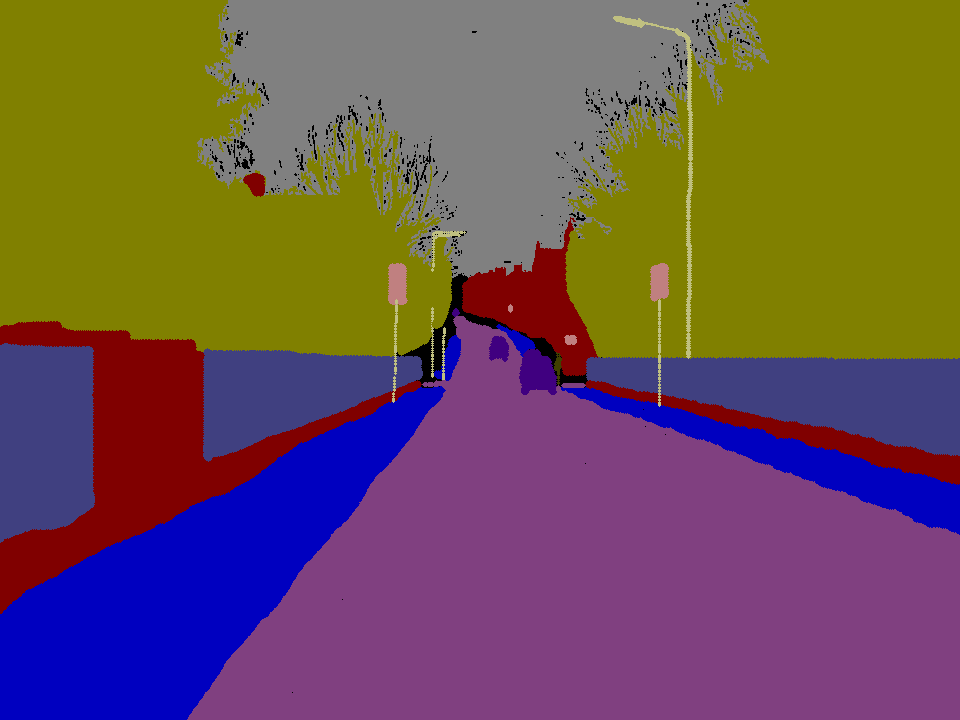}&
\includegraphics[width=.21\textwidth]{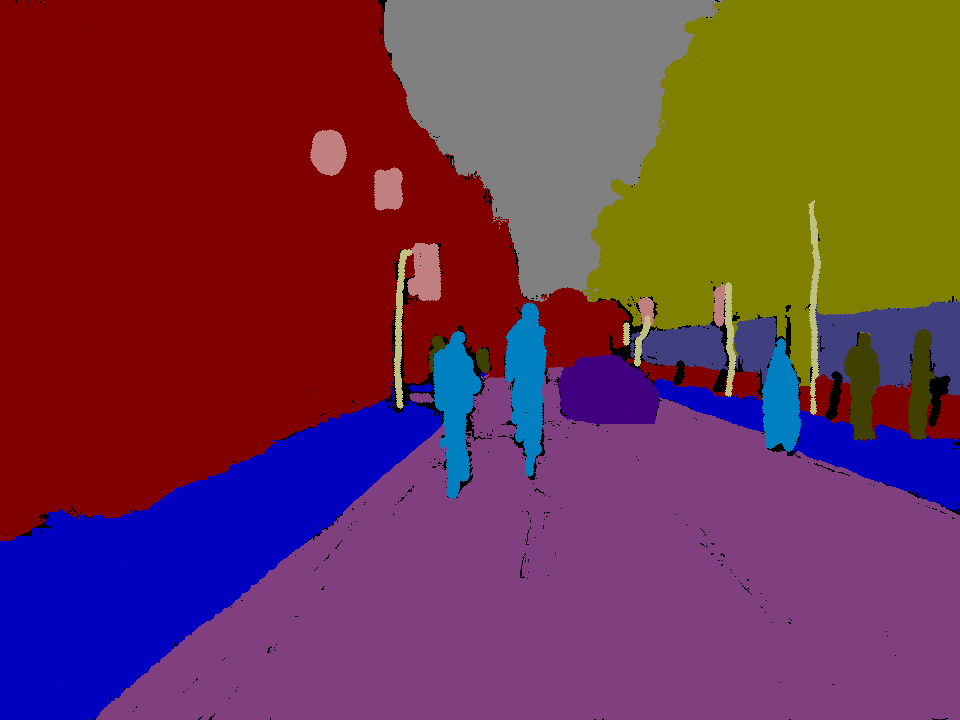}\\
\includegraphics[width=.21\textwidth]{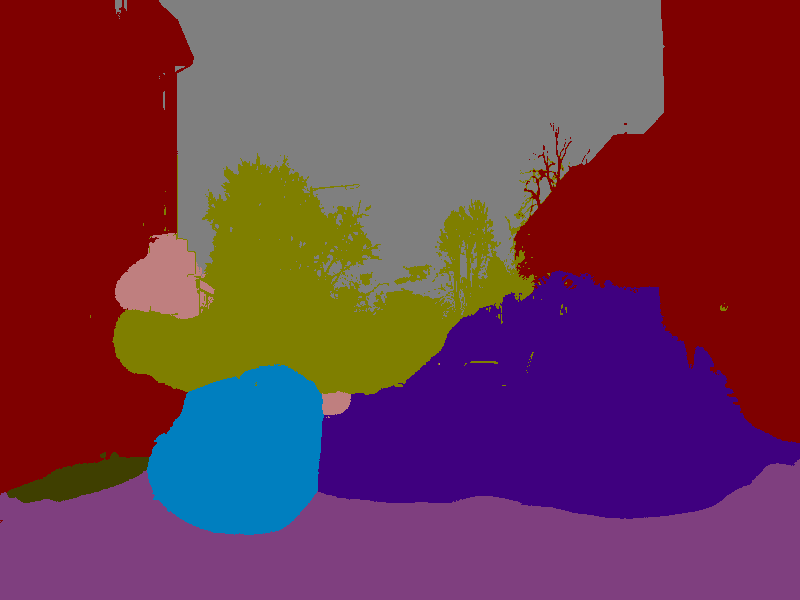}&
\includegraphics[width=.21\textwidth]{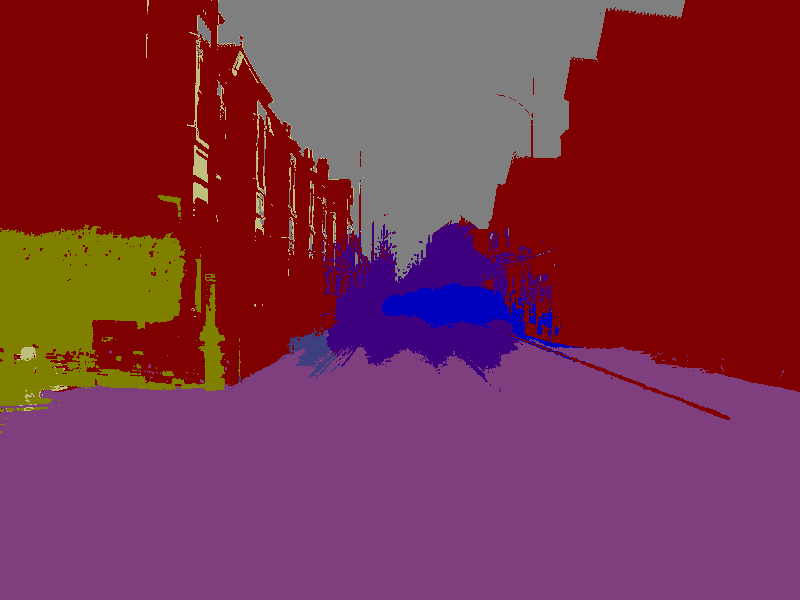}&
\includegraphics[width=.21\textwidth]{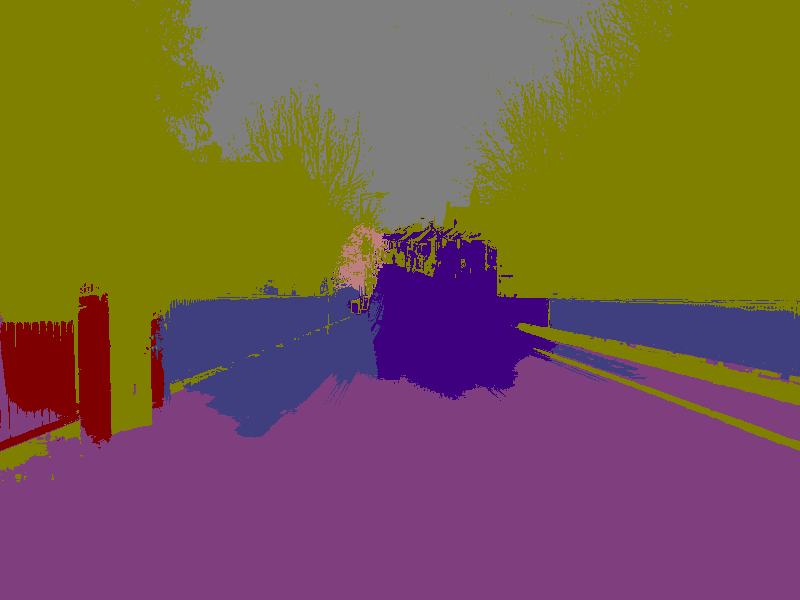}&
\includegraphics[width=.21\textwidth]{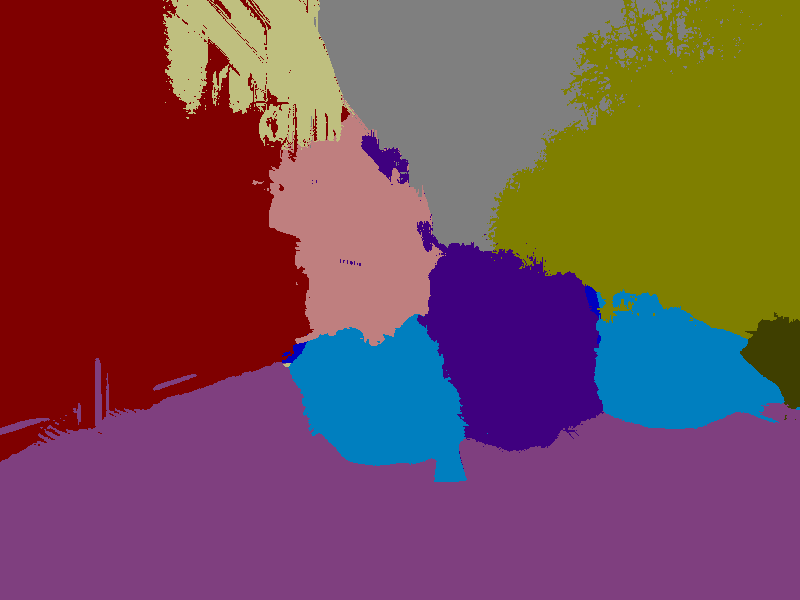}\\
\multicolumn{4}{c}{YouTube-Objects Dataset}\\
\includegraphics[width=.21\textwidth]{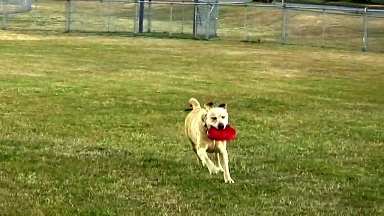}&
\includegraphics[width=.21\textwidth]{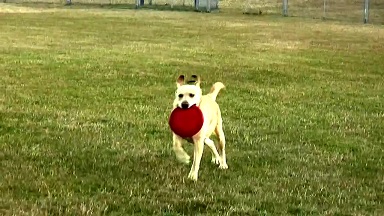}&
\includegraphics[width=.21\textwidth]{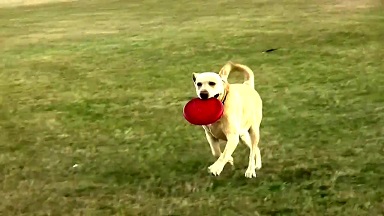}&
\includegraphics[width=.21\textwidth]{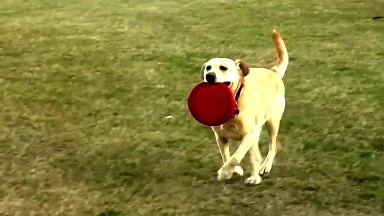}\\
\includegraphics[width=.21\textwidth]{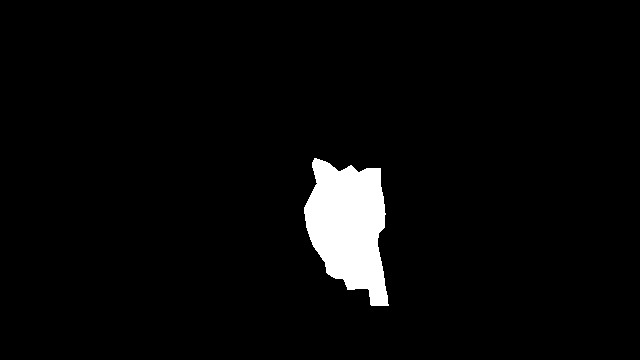}&
\includegraphics[width=.21\textwidth]{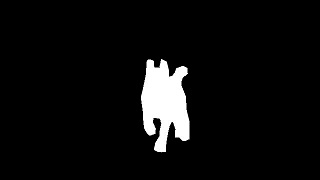}&
\includegraphics[width=.21\textwidth]{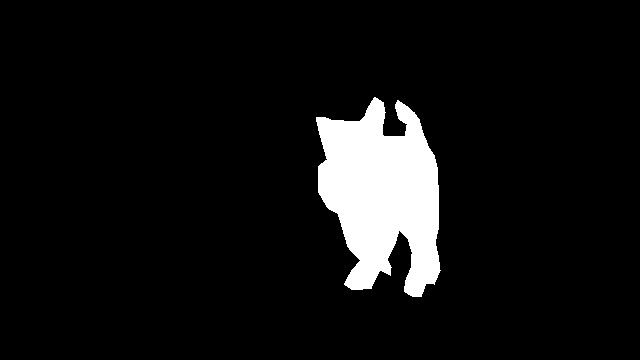}&
\includegraphics[width=.21\textwidth]{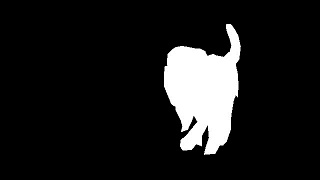}\\
\includegraphics[width=.21\textwidth]{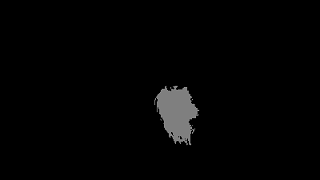}&
\includegraphics[width=.21\textwidth]{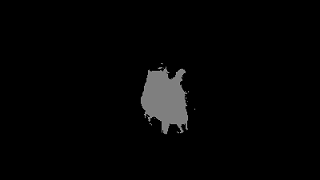}&
\includegraphics[width=.21\textwidth]{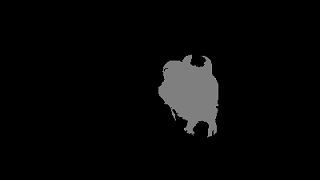}&
\includegraphics[width=.21\textwidth]{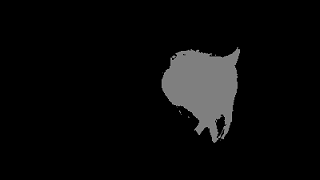}\\
\end{tabular}
\caption{Qualitative results on CityScapes, CamVid, and YouTube-Objects. Note that for each dataset, from top to bottom, there is the RGB frame, Ground-truth and the prediction of our two-stream network.} 
\label{fig:result}
\end{figure*}

To further show the effectiveness of our method, we compare it with other weakly-supervised video semantic segmentation baselines on the standard YouTube-Objects dataset. Note that, here, all the classes correspond to foreground objects, with a single background class, which makes this dataset a less attractive candidate for our method. This comparison, however, lets us evaluate the performance of our two-stream network with respect to the state-of-the-art in weakly-supervised video semantic segmentation. As shown in Table~\ref{tab:yto}, our results significantly outperform the state-of-the-art on this dataset, thus again showing the benefits of our approach (see Fig.~\ref{fig:result} for qualitative results).

Note that other approaches that make use of additional supervision, such as object detectors trained from pixel-level~\cite{viadetection} or bounding box~\cite{obj_track} annotations, have also reported results on this dataset. While we only exploit tags, our approach yields results comparable to those of these methods (53.3\% for our method versus 54.1\% for~\cite{viadetection} and 55.8\% for~\cite{obj_track}).

\section{Conclusion}
In this paper, we have proposed the first weakly-supervised video semantic segmentation approach that considers both multiple foreground and background classes. To this end, we have introduced a two-stream network that leverages optical-flow and RGB image, trained using a loss based on classifier heatmaps. Our experiments have demonstrated the benefits of using such heatmaps and of exploiting optical flow on challenging urban datasets. Furthermore, our two-stream network has also outperformed the state-of-the-art weakly-supervised video semantic segmentation methods on the standard YouTube-Object benchmark. In the future, we plan to investigate other fusion strategies within our two-stream formalism. Moreover, we will aim to leverage depth information from stereo images, which does not require any additional annotations.

{\small
\bibliographystyle{ieee}
\bibliography{egbib}

\begin{thebibliography}{10}\itemsep=-1pt

\bibitem{segnet}
V.~Badrinarayanan, A.~Handa, and R.~Cipolla.
\newblock Segnet: A deep convolutional encoder-decoder architecture for robust
  semantic pixel-wise labelling.
\newblock {\em arXiv preprint arXiv:1505.07293}, 2015.

\bibitem{whatspoint}
A.~Bearman, O.~Russakovsky, V.~Ferrari, and L.~Fei-Fei.
\newblock What’s the point: Semantic segmentation with point supervision.
\newblock In {\em European Conference on Computer Vision}, pages 549--565.
  Springer, 2016.

\bibitem{deepedge}
G.~Bertasius, J.~Shi, and L.~Torresani.
\newblock Deepedge: A multi-scale bifurcated deep network for top-down contour
  detection.
\newblock In {\em Proceedings of the IEEE Conference on Computer Vision and
  Pattern Recognition}, pages 4380--4389, 2015.

\bibitem{camvid}
G.~J. Brostow, J.~Fauqueur, and R.~Cipolla.
\newblock Semantic object classes in video: A high-definition ground truth
  database.
\newblock {\em Pattern Recognition Letters}, 30(2):88--97, 2009.

\bibitem{camvid11}
G.~J. Brostow, J.~Fauqueur, and R.~Cipolla.
\newblock Semantic object classes in video: A high-definition ground truth
  database.
\newblock {\em Pattern Recognition Letters}, 30(2):88--97, 2009.

\bibitem{optical-flow}
T.~Brox, A.~Bruhn, N.~Papenberg, and J.~Weickert.
\newblock High accuracy optical flow estimation based on a theory for warping.
\newblock In {\em European conference on computer vision}, pages 25--36.
  Springer, 2004.

\bibitem{flow-yto}
T.~Brox and J.~Malik.
\newblock Large displacement optical flow: descriptor matching in variational
  motion estimation.
\newblock {\em IEEE transactions on pattern analysis and machine intelligence},
  33(3):500--513, 2011.

\bibitem{cpmc}
J.~Carreira and C.~Sminchisescu.
\newblock Cpmc: Automatic object segmentation using constrained parametric
  min-cuts.
\newblock {\em IEEE Transactions on Pattern Analysis and Machine Intelligence},
  34(7):1312--1328, 2012.

\bibitem{deeplab}
L.-C. Chen, G.~Papandreou, I.~Kokkinos, K.~Murphy, and A.~L. Yuille.
\newblock Semantic image segmentation with deep convolutional nets and fully
  connected crfs.
\newblock {\em arXiv preprint arXiv:1412.7062}, 2014.

\bibitem{Cityscapes}
M.~Cordts, M.~Omran, S.~Ramos, T.~Rehfeld, M.~Enzweiler, R.~Benenson,
  U.~Franke, S.~Roth, and B.~Schiele.
\newblock The cityscapes dataset for semantic urban scene understanding.
\newblock In {\em Proc. of the IEEE Conference on Computer Vision and Pattern
  Recognition (CVPR)}, 2016.

\bibitem{inria}
N.~Dalal and B.~Triggs.
\newblock Histograms of oriented gradients for human detection.
\newblock In {\em Computer Vision and Pattern Recognition, 2005. CVPR 2005.
  IEEE Computer Society Conference on}, volume~1, pages 886--893. IEEE, 2005.

\bibitem{image-net}
J.~Deng, W.~Dong, R.~Socher, L.-J. Li, K.~Li, and L.~Fei-Fei.
\newblock Imagenet: A large-scale hierarchical image database.
\newblock \url{http://image-net.org}, 2009.

\bibitem{obj_track}
B.~Drayer and T.~Brox.
\newblock Object detection, tracking, and motion segmentation for object-level
  video segmentation.
\newblock {\em arXiv preprint arXiv:1608.03066}, 2016.

\bibitem{2stream}
C.~Feichtenhofer, A.~Pinz, and A.~Zisserman.
\newblock Convolutional two-stream network fusion for video action recognition.
\newblock In {\em Proceedings of the IEEE Conference on Computer Vision and
  Pattern Recognition}, pages 1933--1941, 2016.

\bibitem{movingobject}
K.~Fragkiadaki, P.~Arbelaez, P.~Felsen, and J.~Malik.
\newblock Learning to segment moving objects in videos.
\newblock In {\em Proceedings of the IEEE Conference on Computer Vision and
  Pattern Recognition}, pages 4083--4090, 2015.

\bibitem{webscale}
G.~Hartmann, M.~Grundmann, J.~Hoffman, D.~Tsai, V.~Kwatra, O.~Madani,
  S.~Vijayanarasimhan, I.~Essa, J.~Rehg, and R.~Sukthankar.
\newblock Weakly supervised learning of object segmentations from web-scale
  video.
\newblock In {\em European Conference on Computer Vision}, pages 198--208.
  Springer, 2012.

\bibitem{webcrawled}
S.~Hong, D.~Yeo, S.~Kwak, H.~Lee, and B.~Han.
\newblock Weakly supervised semantic segmentation using web-crawled videos.
\newblock {\em arXiv preprint arXiv:1701.00352}, 2017.

\bibitem{mining}
Q.~Hou, P.~K. Dokania, D.~Massiceti, Y.~Wei, M.-M. Cheng, and P.~Torr.
\newblock Mining pixels: Weakly supervised semantic segmentation using image
  labels.
\newblock {\em arXiv preprint arXiv:1612.02101}, 2016.

\bibitem{supervoxel}
S.~D. Jain and K.~Grauman.
\newblock Supervoxel-consistent foreground propagation in video.
\newblock In {\em European Conference on Computer Vision}, pages 656--671.
  Springer, 2014.

\bibitem{mask-yto}
S.~D. Jain and K.~Grauman.
\newblock Supervoxel-consistent foreground propagation in video.
\newblock In {\em European Conference on Computer Vision}, pages 656--671.
  Springer, 2014.

\bibitem{caffe}
Y.~Jia, E.~Shelhamer, J.~Donahue, S.~Karayev, J.~Long, R.~Girshick,
  S.~Guadarrama, and T.~Darrell.
\newblock Caffe: Convolutional architecture for fast feature embedding.
\newblock {\em arXiv preprint arXiv:1408.5093}, 2014.

\bibitem{predective}
X.~Jin, X.~Li, H.~Xiao, X.~Shen, Z.~Lin, J.~Yang, Y.~Chen, J.~Dong, L.~Liu,
  Z.~Jie, et~al.
\newblock Video scene parsing with predictive feature learning.
\newblock {\em arXiv preprint arXiv:1612.00119}, 2016.

\bibitem{sec}
A.~Kolesnikov and C.~H. Lampert.
\newblock Seed, expand and constrain: Three principles for weakly-supervised
  image segmentation.
\newblock In {\em European Conference on Computer Vision}, pages 695--711.
  Springer, 2016.

\bibitem{dcrf}
P.~Kr{\"a}henb{\"u}hl and V.~Koltun.
\newblock Efficient inference in fully connected crfs with gaussian edge
  potentials.
\newblock In {\em Advances in Neural Information Processing Systems}, pages
  109--117, 2011.

\bibitem{featurespace}
A.~Kundu, V.~Vineet, and V.~Koltun.
\newblock Feature space optimization for semantic video segmentation.
\newblock In {\em Proceedings of the IEEE Conference on Computer Vision and
  Pattern Recognition}, pages 3168--3175, 2016.

\bibitem{activeinference}
B.~Liu and X.~He.
\newblock Multiclass semantic video segmentation with object-level active
  inference.
\newblock In {\em Proceedings of the IEEE Conference on Computer Vision and
  Pattern Recognition}, pages 4286--4294, 2015.

\bibitem{multi-class}
X.~Liu, D.~Tao, M.~Song, Y.~Ruan, C.~Chen, and J.~Bu.
\newblock Weakly supervised multiclass video segmentation.
\newblock In {\em Proceedings of the IEEE Conference on Computer Vision and
  Pattern Recognition}, pages 57--64, 2014.

\bibitem{FCN}
J.~Long, E.~Shelhamer, and T.~Darrell.
\newblock Fully convolutional networks for semantic segmentation.
\newblock In {\em Proceedings of the IEEE Conference on Computer Vision and
  Pattern Recognition}, pages 3431--3440, 2015.

\bibitem{diverse}
M.~Mostajabi, N.~Kolkin, and G.~Shakhnarovich.
\newblock Diverse sampling for self-supervised learning of semantic
  segmentation.
\newblock {\em arXiv preprint arXiv:1612.01991}, 2016.

\bibitem{long-term-video}
P.~Ochs, J.~Malik, and T.~Brox.
\newblock Segmentation of moving objects by long term video analysis.
\newblock {\em IEEE transactions on pattern analysis and machine intelligence},
  36(6):1187--1200, 2014.

\bibitem{exploiting}
S.~J. Oh, R.~Benenson, A.~Khoreva, Z.~Akata, M.~Fritz, and B.~Schiele.
\newblock Exploiting saliency for object segmentation from image level labels.
\newblock {\em arXiv preprint arXiv:1701.08261}, 2017.

\bibitem{oquab2015object}
M.~Oquab, L.~Bottou, I.~Laptev, and J.~Sivic.
\newblock Is object localization for free?-weakly-supervised learning with
  convolutional neural networks.
\newblock In {\em Proceedings of the IEEE Conference on Computer Vision and
  Pattern Recognition}, pages 685--694, 2015.

\bibitem{nicta-ped}
G.~Overett, L.~Petersson, N.~Brewer, L.~Andersson, and N.~Pettersson.
\newblock A new pedestrian dataset for supervised learning.
\newblock In {\em Intelligent Vehicles Symposium, 2008 IEEE}, pages 373--378.
  IEEE, 2008.

\bibitem{weaklyandsemi}
G.~Papandreou, L.-C. Chen, K.~P. Murphy, and A.~L. Yuille.
\newblock Weakly-and semi-supervised learning of a deep convolutional network
  for semantic image segmentation.
\newblock In {\em Proceedings of the IEEE International Conference on Computer
  Vision}, pages 1742--1750, 2015.

\bibitem{fast}
A.~Papazoglou and V.~Ferrari.
\newblock Fast object segmentation in unconstrained video.
\newblock In {\em Proceedings of the IEEE International Conference on Computer
  Vision}, pages 1777--1784, 2013.

\bibitem{ccnn}
D.~Pathak, P.~Krahenbuhl, and T.~Darrell.
\newblock Constrained convolutional neural networks for weakly supervised
  segmentation.
\newblock In {\em Proceedings of the IEEE International Conference on Computer
  Vision}, pages 1796--1804, 2015.

\bibitem{mil}
D.~Pathak, E.~Shelhamer, J.~Long, and T.~Darrell.
\newblock Fully convolutional multi-class multiple instance learning.
\newblock {\em arXiv preprint arXiv:1412.7144}, 2014.

\bibitem{fromimagelevel}
P.~O. Pinheiro and R.~Collobert.
\newblock From image-level to pixel-level labeling with convolutional networks.
\newblock In {\em Proceedings of the IEEE Conference on Computer Vision and
  Pattern Recognition}, pages 1713--1721, 2015.

\bibitem{graph-based}
N.~Pourian, S.~Karthikeyan, and B.~Manjunath.
\newblock Weakly supervised graph based semantic segmentation by learning
  communities of image-parts.
\newblock In {\em Proceedings of the IEEE International Conference on Computer
  Vision}, pages 1359--1367, 2015.

\bibitem{youtube}
A.~Prest, C.~Leistner, J.~Civera, C.~Schmid, and V.~Ferrari.
\newblock Learning object class detectors from weakly annotated video.
\newblock In {\em Computer Vision and Pattern Recognition (CVPR), 2012 IEEE
  Conference on}, pages 3282--3289. IEEE, 2012.

\bibitem{augmented}
X.~Qi, Z.~Liu, J.~Shi, H.~Zhao, and J.~Jia.
\newblock Augmented feedback in semantic segmentation under image level
  supervision.
\newblock In {\em European Conference on Computer Vision}, pages 90--105.
  Springer, 2016.

\bibitem{playgame}
S.~R. Richter, V.~Vineet, S.~Roth, and V.~Koltun.
\newblock Playing for data: Ground truth from computer games.
\newblock In {\em European Conference on Computer Vision}, pages 102--118.
  Springer, 2016.

\bibitem{synthia}
G.~Ros, L.~Sellart, J.~Materzynska, D.~Vazquez, and A.~M. Lopez.
\newblock The synthia dataset: A large collection of synthetic images for
  semantic segmentation of urban scenes.
\newblock In {\em Proceedings of the IEEE Conference on Computer Vision and
  Pattern Recognition}, pages 3234--3243, 2016.

\bibitem{ILSVRC}
O.~Russakovsky, J.~Deng, H.~Su, J.~Krause, S.~Satheesh, S.~Ma, Z.~Huang,
  A.~Karpathy, A.~Khosla, M.~Bernstein, A.~C. Berg, and L.~Fei-Fei.
\newblock {ImageNet Large Scale Visual Recognition Challenge}.
\newblock {\em International Journal of Computer Vision (IJCV)},
  115(3):211--252, 2015.

\bibitem{saleh}
F.~Saleh, M.~S.~A. Akbarian, M.~Salzmann, L.~Petersson, S.~Gould, and J.~M.
  Alvarez.
\newblock Built-in foreground/background prior for weakly-supervised semantic
  segmentation.
\newblock In {\em European Conference on Computer Vision}, pages 413--432.
  Springer, 2016.

\bibitem{viastroke}
N.~Shankar~Nagaraja, F.~R. Schmidt, and T.~Brox.
\newblock Video segmentation with just a few strokes.
\newblock In {\em Proceedings of the IEEE International Conference on Computer
  Vision}, pages 3235--3243, 2015.

\bibitem{clockwork}
E.~Shelhamer, K.~Rakelly, J.~Hoffman, and T.~Darrell.
\newblock Clockwork convnets for video semantic segmentation.
\newblock In {\em Computer Vision--ECCV 2016 Workshops}, pages 852--868.
  Springer, 2016.

\bibitem{distinct_saliency}
W.~Shimoda and K.~Yanai.
\newblock Distinct class-specific saliency maps for weakly supervised semantic
  segmentation.
\newblock In {\em European Conference on Computer Vision}, pages 218--234.
  Springer, 2016.

\bibitem{vgg}
K.~Simonyan and A.~Zisserman.
\newblock Very deep convolutional networks for large-scale image recognition.
\newblock {\em arXiv preprint arXiv:1409.1556}, 2014.

\bibitem{discriminative}
K.~Tang, R.~Sukthankar, J.~Yagnik, and L.~Fei-Fei.
\newblock Discriminative segment annotation in weakly labeled video.
\newblock In {\em Proceedings of the IEEE conference on computer vision and
  pattern recognition}, pages 2483--2490, 2013.

\bibitem{motioncues}
P.~Tokmakov, K.~Alahari, and C.~Schmid.
\newblock Weakly-supervised semantic segmentation using motion cues.
\newblock In {\em European Conference on Computer Vision}, pages 388--404.
  Springer, 2016.

\bibitem{voxel2voxel}
D.~Tran, L.~Bourdev, R.~Fergus, L.~Torresani, and M.~Paluri.
\newblock Deep end2end voxel2voxel prediction.
\newblock In {\em Proceedings of the IEEE Conference on Computer Vision and
  Pattern Recognition Workshops}, pages 17--24, 2016.

\bibitem{tripathi}
S.~Tripathi, S.~Belongie, Y.~Hwang, and T.~Nguyen.
\newblock Semantic video segmentation: Exploring inference efficiency.
\newblock In {\em SoC Design Conference (ISOCC), 2015 International}, pages
  157--158. IEEE, 2015.

\bibitem{viaflow}
Y.-H. Tsai, M.-H. Yang, and M.~J. Black.
\newblock Video segmentation via object flow.
\newblock In {\em Proceedings of the IEEE Conference on Computer Vision and
  Pattern Recognition}, pages 3899--3908, 2016.

\bibitem{co_segment}
Y.-H. Tsai, G.~Zhong, and M.-H. Yang.
\newblock Semantic co-segmentation in videos.
\newblock In {\em European Conference on Computer Vision}, pages 760--775.
  Springer, 2016.

\bibitem{multi-image-model}
A.~Vezhnevets, V.~Ferrari, and J.~M. Buhmann.
\newblock Weakly supervised semantic segmentation with a multi-image model.
\newblock In {\em Computer Vision (ICCV), 2011 IEEE International Conference
  on}, pages 643--650. IEEE, 2011.

\bibitem{structured}
A.~Vezhnevets, V.~Ferrari, and J.~M. Buhmann.
\newblock Weakly supervised structured output learning for semantic
  segmentation.
\newblock In {\em Computer Vision and Pattern Recognition (CVPR), 2012 IEEE
  Conference on}, pages 845--852. IEEE, 2012.

\bibitem{domain_adapt}
H.~Wang, T.~Raiko, L.~Lensu, T.~Wang, and J.~Karhunen.
\newblock Semi-supervised domain adaptation for weakly labeled semantic video
  object segmentation.
\newblock {\em arXiv preprint arXiv:1606.02280}, 2016.

\bibitem{learning}
Y.~Wei, X.~Liang, Y.~Chen, Z.~Jie, Y.~Xiao, Y.~Zhao, and S.~Yan.
\newblock Learning to segment with image-level annotations.
\newblock {\em Pattern Recognition}, 59:234--244, 2016.

\bibitem{stc}
Y.~Wei, X.~Liang, Y.~Chen, X.~Shen, M.-M. Cheng, J.~Feng, Y.~Zhao, and S.~Yan.
\newblock Stc: A simple to complex framework for weakly-supervised semantic
  segmentation.
\newblock {\em IEEE Transactions on Pattern Analysis and Machine Intelligence},
  2016.

\bibitem{tellme}
J.~Xu, A.~G. Schwing, and R.~Urtasun.
\newblock Tell me what you see and i will show you where it is.
\newblock In {\em Proceedings of the IEEE Conference on Computer Vision and
  Pattern Recognition}, pages 3190--3197, 2014.

\bibitem{social}
W.~Zhang, S.~Zeng, D.~Wang, and X.~Xue.
\newblock Weakly supervised semantic segmentation for social images.
\newblock In {\em Proceedings of the IEEE Conference on Computer Vision and
  Pattern Recognition}, pages 2718--2726, 2015.

\bibitem{viadetection}
Y.~Zhang, X.~Chen, J.~Li, C.~Wang, and C.~Xia.
\newblock Semantic object segmentation via detection in weakly labeled video.
\newblock In {\em Proceedings of the IEEE Conference on Computer Vision and
  Pattern Recognition}, pages 3641--3649, 2015.

\bibitem{co_parsing}
G.~Zhong12, Y.-H. Tsai, and M.-H. Yang.
\newblock Weakly-supervised video scene co-parsing.

\bibitem{zhou2016learning}
B.~Zhou, A.~Khosla, A.~Lapedriza, A.~Oliva, and A.~Torralba.
\newblock Learning deep features for discriminative localization.
\newblock In {\em Proceedings of the IEEE Conference on Computer Vision and
  Pattern Recognition}, pages 2921--2929, 2016.

\end{thebibliography}
}


\end{document}